\documentclass[pdflatex,sn-mathphys-num]{sn-jnl}

\usepackage{graphicx}%
\usepackage{multirow}%
\usepackage{amsmath,amssymb,amsfonts}%
\usepackage{amsthm}%
\usepackage{mathrsfs}%
\usepackage[title]{appendix}%
\usepackage{xcolor}%
\usepackage{textcomp}%
\usepackage{manyfoot}%
\usepackage{booktabs}%
\usepackage{algorithm}%
\usepackage{algorithmicx}%
\usepackage{algpseudocode}%
\usepackage{listings}%
\usepackage{graphicx}
\usepackage{makecell}
\usepackage[final]{changes}
\usepackage{subcaption}

\theoremstyle{thmstyleone}%
\theoremstyle{thmstyletwo}%

\theoremstyle{thmstylethree}%

\begin{document}

\title[Article Title]{Fast Sampling Through The Reuse Of Attention Maps In Diffusion Models}

\author*[1]{\fnm{Rosco} \sur{Hunter}}
\author*[2]{\fnm{{\L}ukasz} \sur{Dudziak}}
\author[3]{\fnm{Mohamed S.} \sur{Abdelfattah}}
\author[2]{\fnm{Abhinav} \sur{Mehrotra}}
\author[2]{\fnm{Sourav} \sur{Bhattacharya}}
\author*[1,2]{\fnm{Hongkai} \sur{Wen}}

\affil[1]{\orgname{University of Warwick},\country{UK}}
\affil[2]{\orgname{Samsung AI Centre Cambridge},\country{UK}}
\affil[3]{\orgname{Cornell University},\country{USA}}

\abstract{Text-to-image diffusion models have demonstrated unprecedented capabilities for flexible and realistic image synthesis.
    Nevertheless, these models rely on a time-consuming sampling procedure, which has motivated attempts to reduce their latency.
    When improving efficiency, researchers often use the original diffusion model to train an additional network designed specifically for fast image generation.
    In contrast, our approach seeks to reduce latency directly, without any retraining, fine-tuning, or knowledge distillation.    
    In particular, we find the repeated calculation of attention maps to be costly yet redundant, and instead suggest reusing them during sampling.
    Our specific reuse strategies are based on ODE theory, which implies that the later a map is reused, the smaller the distortion in the final image. 
    We empirically compare \deleted{these}\added{our} reuse strategies with few-step sampling procedures of comparable latency, finding that reuse generates images that are closer to those produced by the original high-latency diffusion model.\textcolor{white}{\footnote{Rosco Hunter and {\L}ukasz Dudziak contributed equally. \\ \textcolor{white}{*}Hongkai Wen is the corresponding author.}}}


\maketitle

\begin{figure*}
\begin{minipage}[b]{0.9\textwidth}
\setlength{\tabcolsep}{3pt}
\begin{tabular}{ccccccc}
\small
\raisebox{0.9cm}{\makecell[tt]{\textbf{DDIM} \\ \textit{20-step}}} & 
\includegraphics[width=0.13\textwidth]{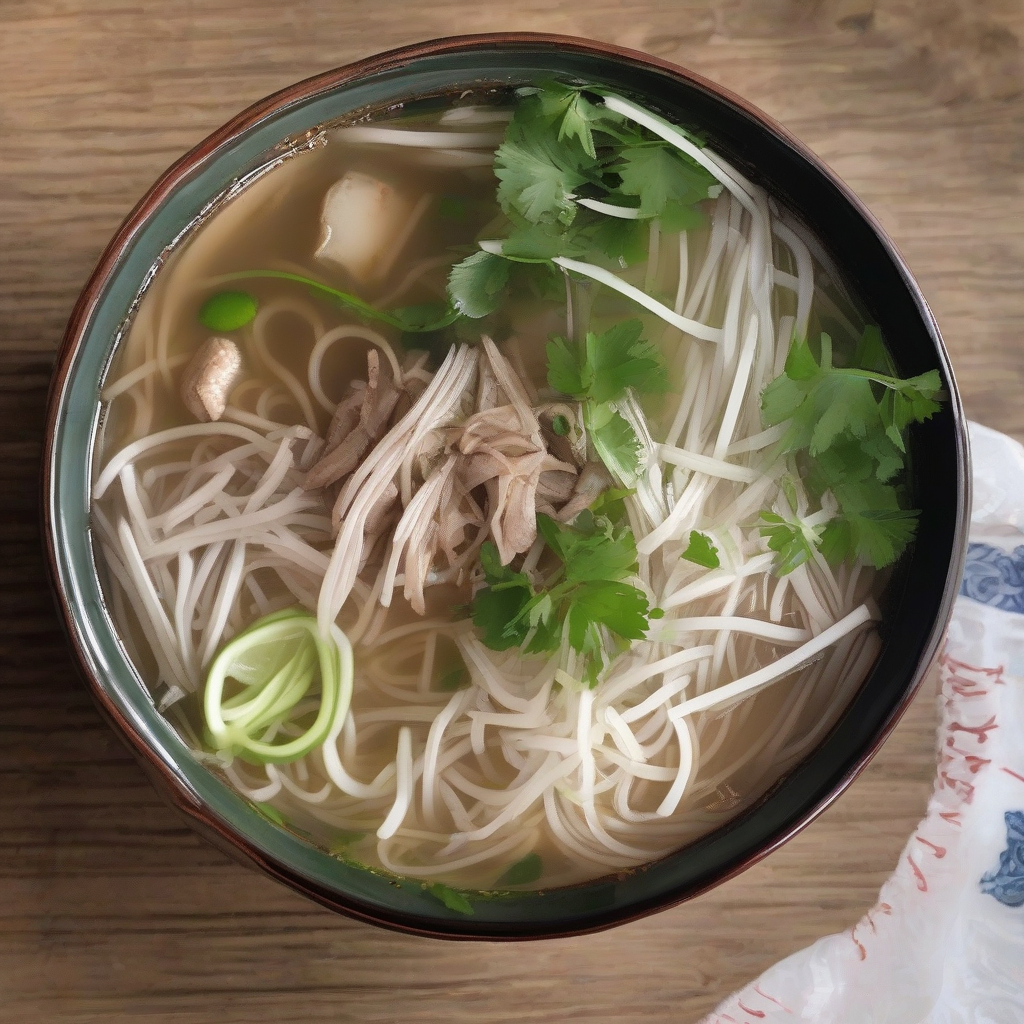} & \includegraphics[width=0.13\textwidth]{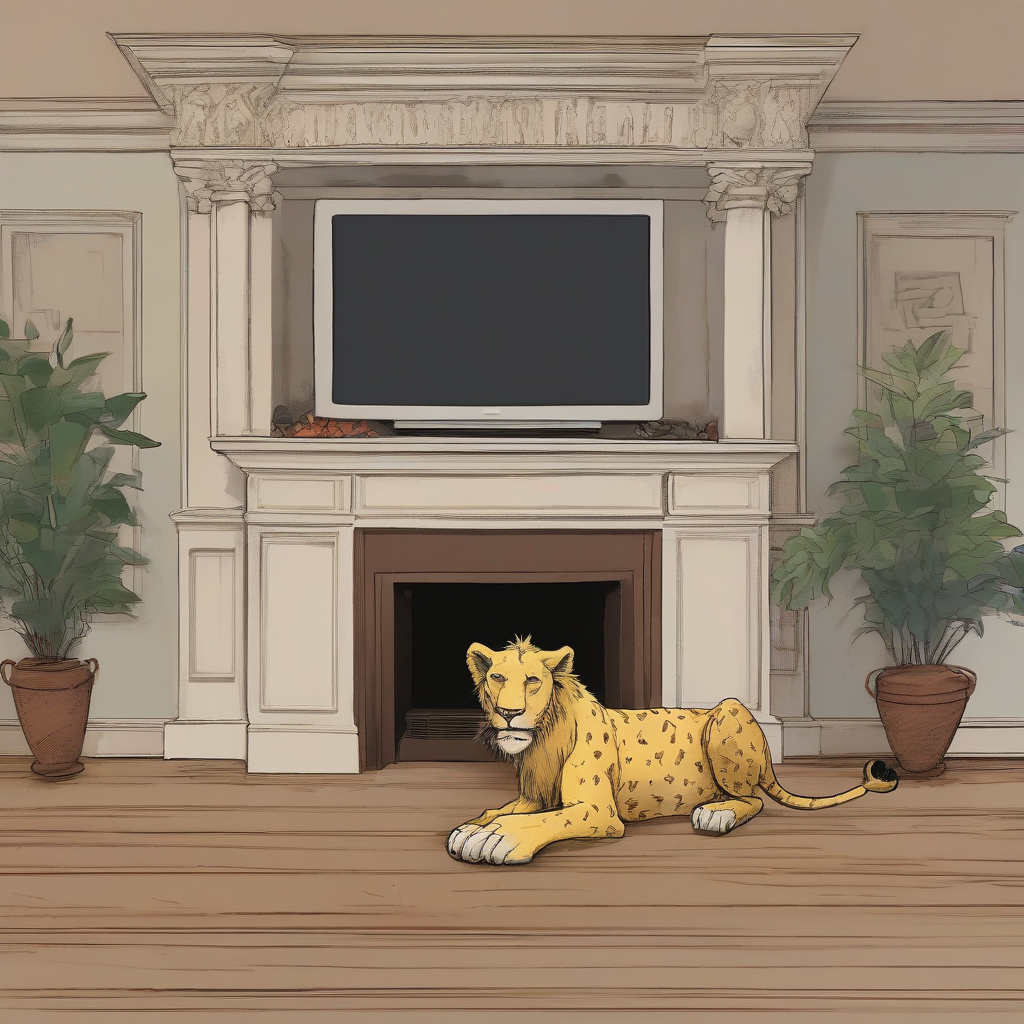} & \includegraphics[width=0.13\textwidth]{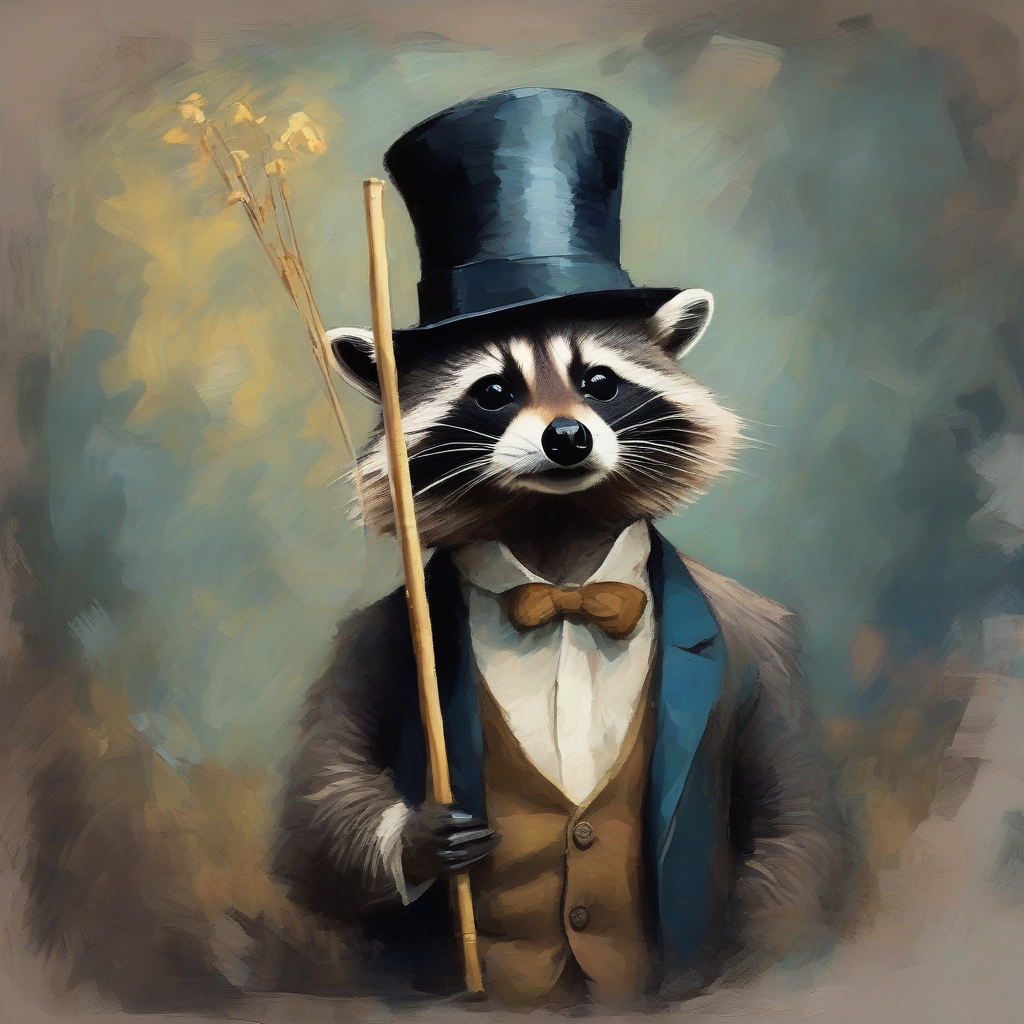} & \includegraphics[width=0.13\textwidth]{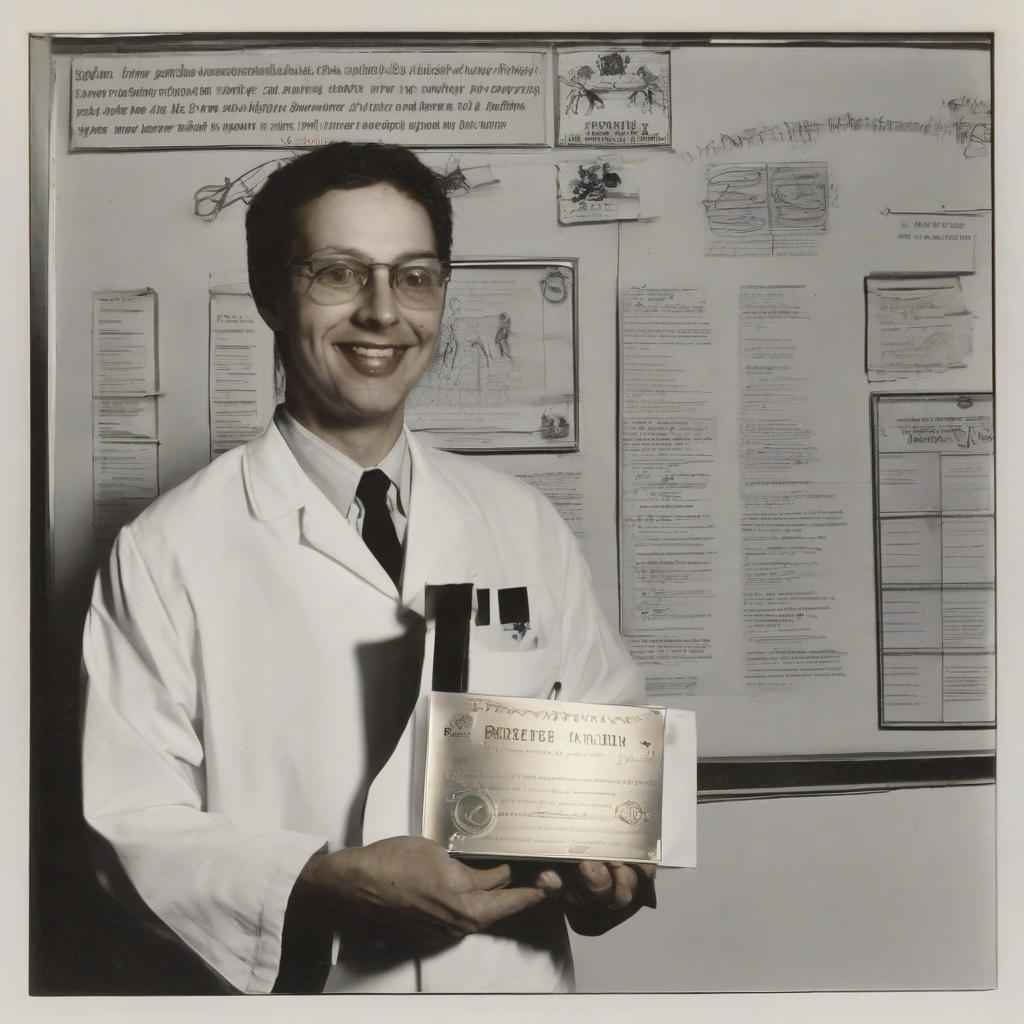} & \includegraphics[width=0.13\textwidth]{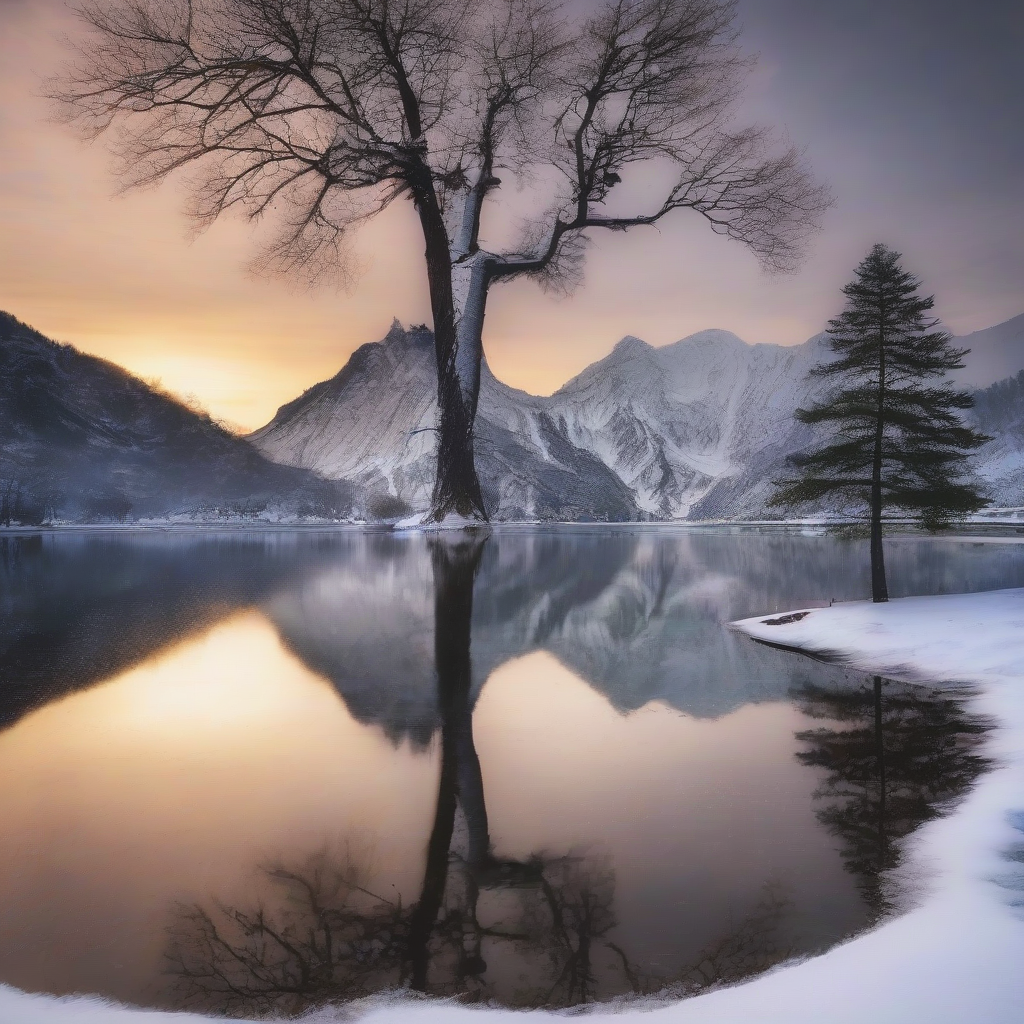} & \includegraphics[width=0.13\textwidth]{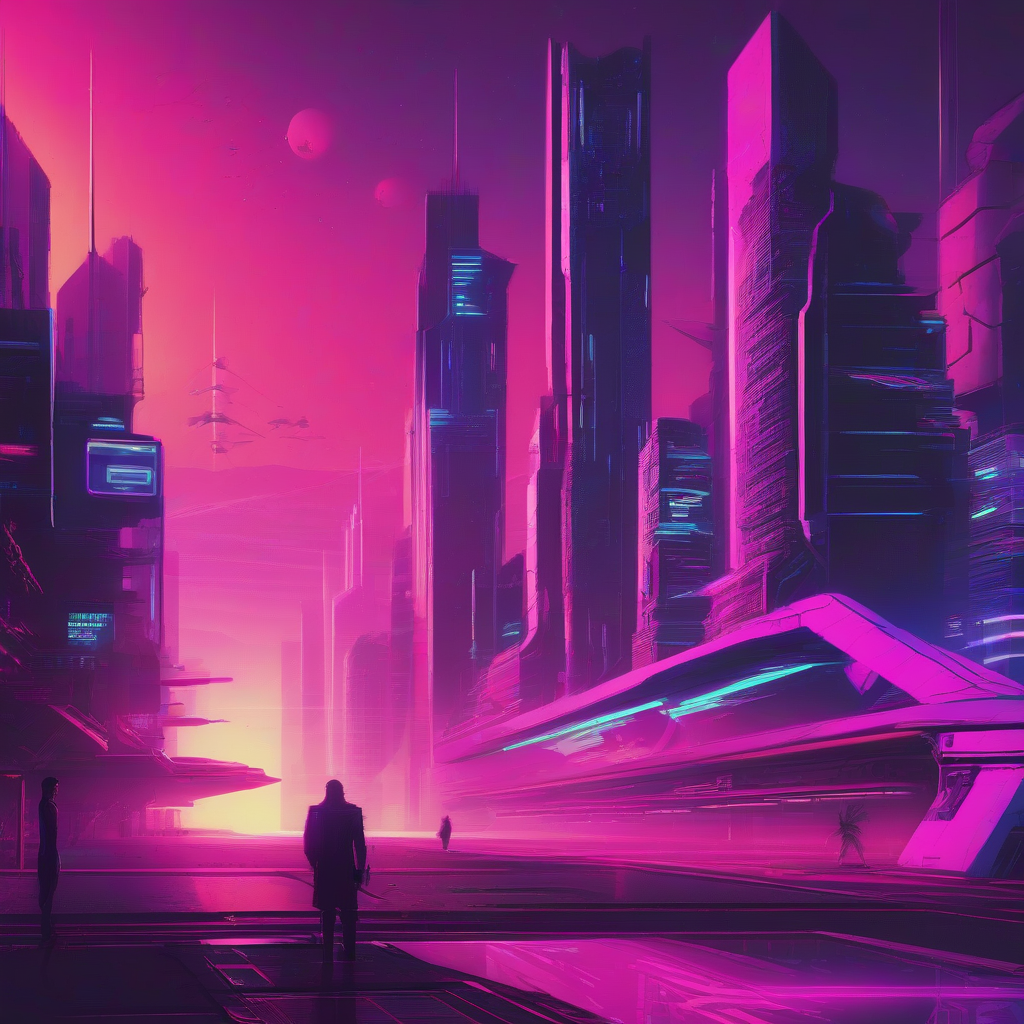} \\
\raisebox{1.1cm}{\makecell[tt]{\textbf{DDIM} \\  \textit{20-step} with \\  \textit{10 reuse steps}}}  & 
\includegraphics[width=0.13\textwidth]{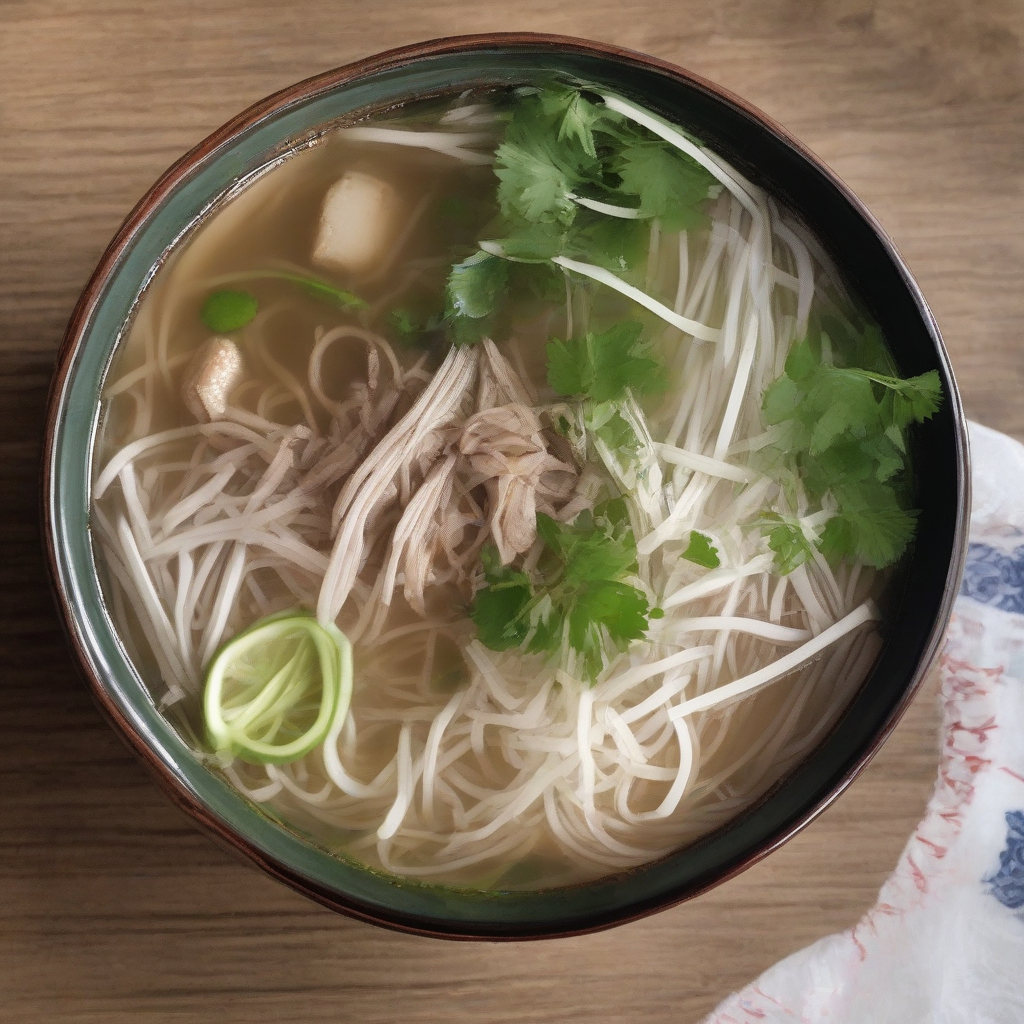} & \includegraphics[width=0.13\textwidth]{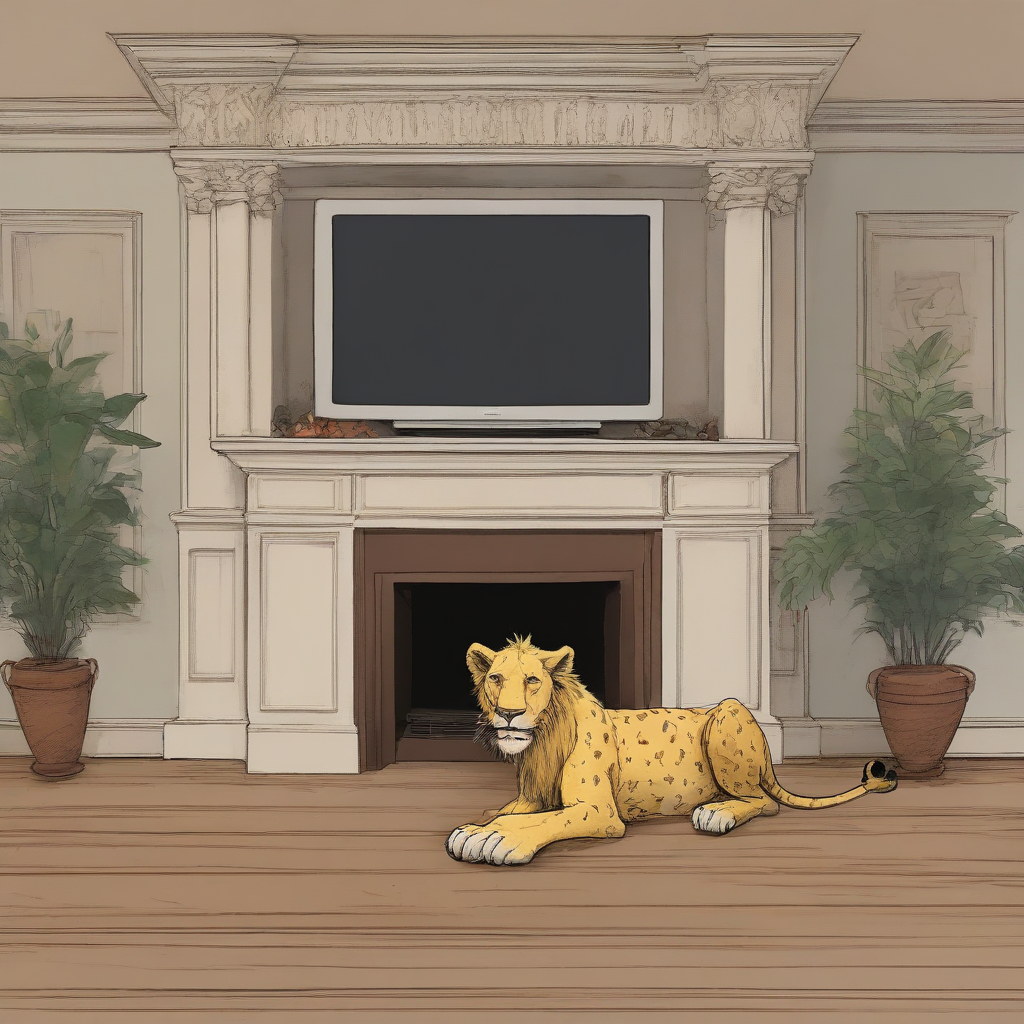} &  \includegraphics[width=0.13\textwidth]{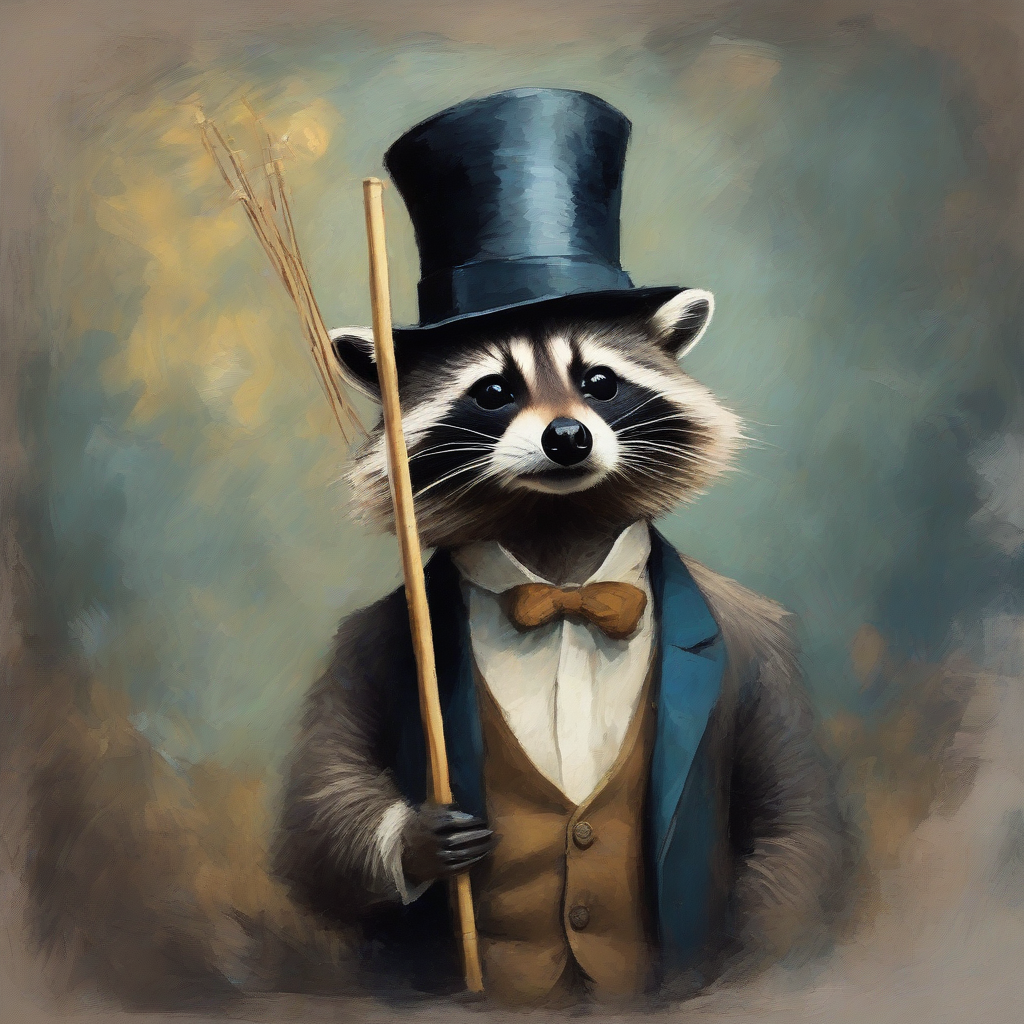} & \includegraphics[width=0.13\textwidth]{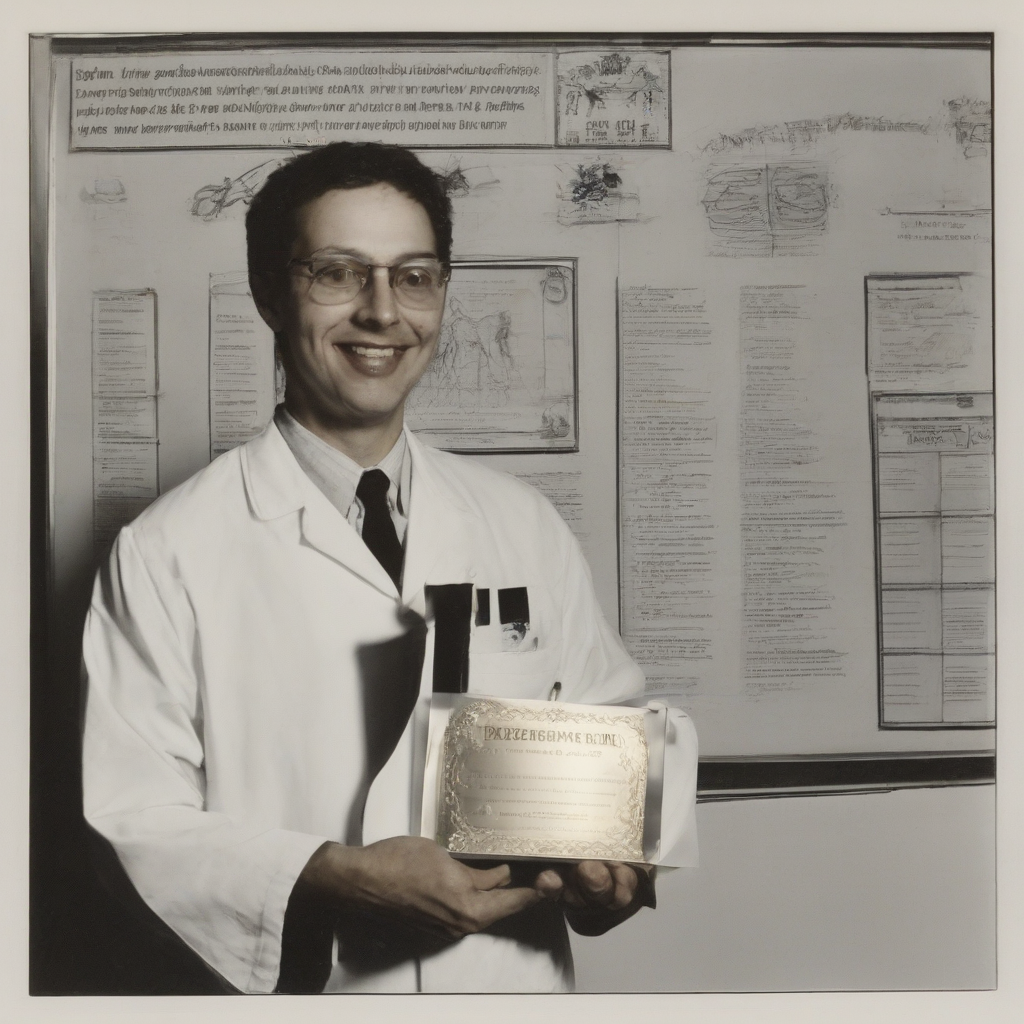} & \includegraphics[width=0.13\textwidth]{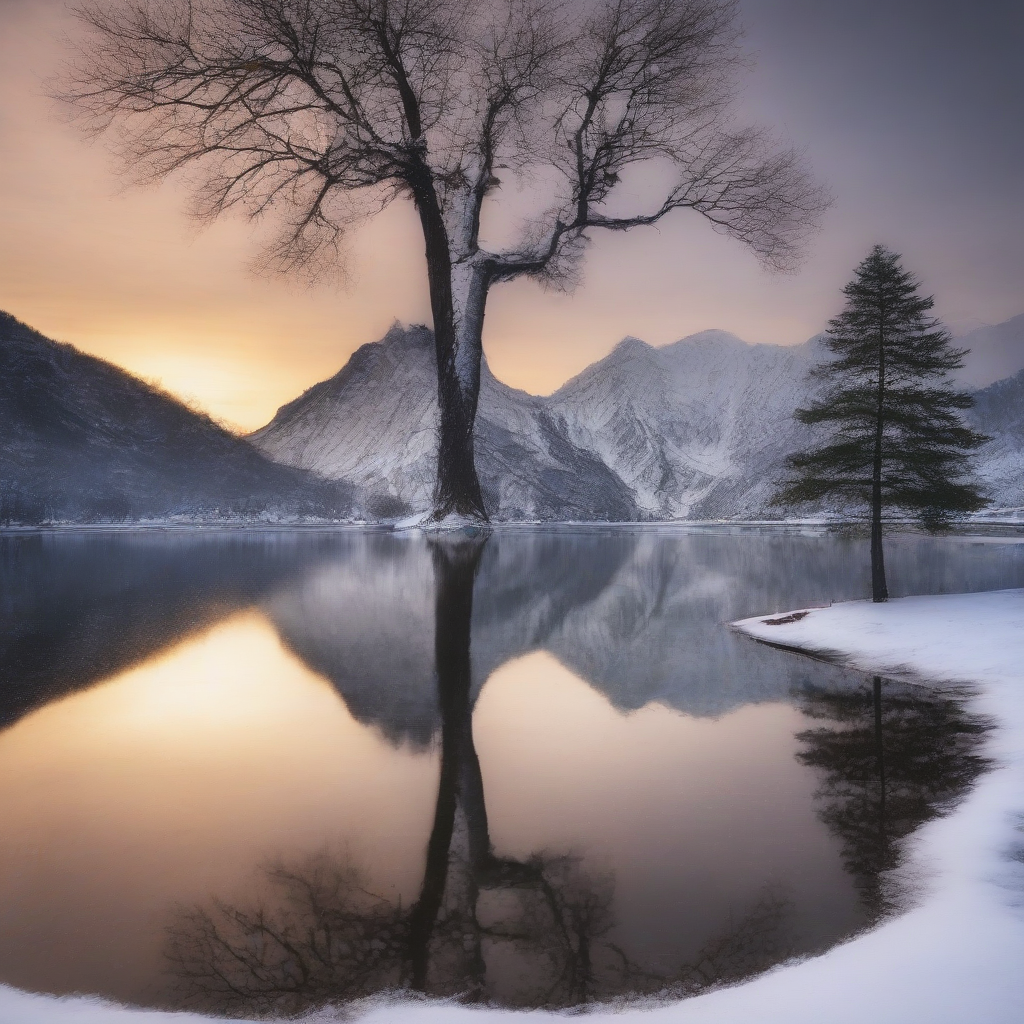} & \includegraphics[width=0.13\textwidth]{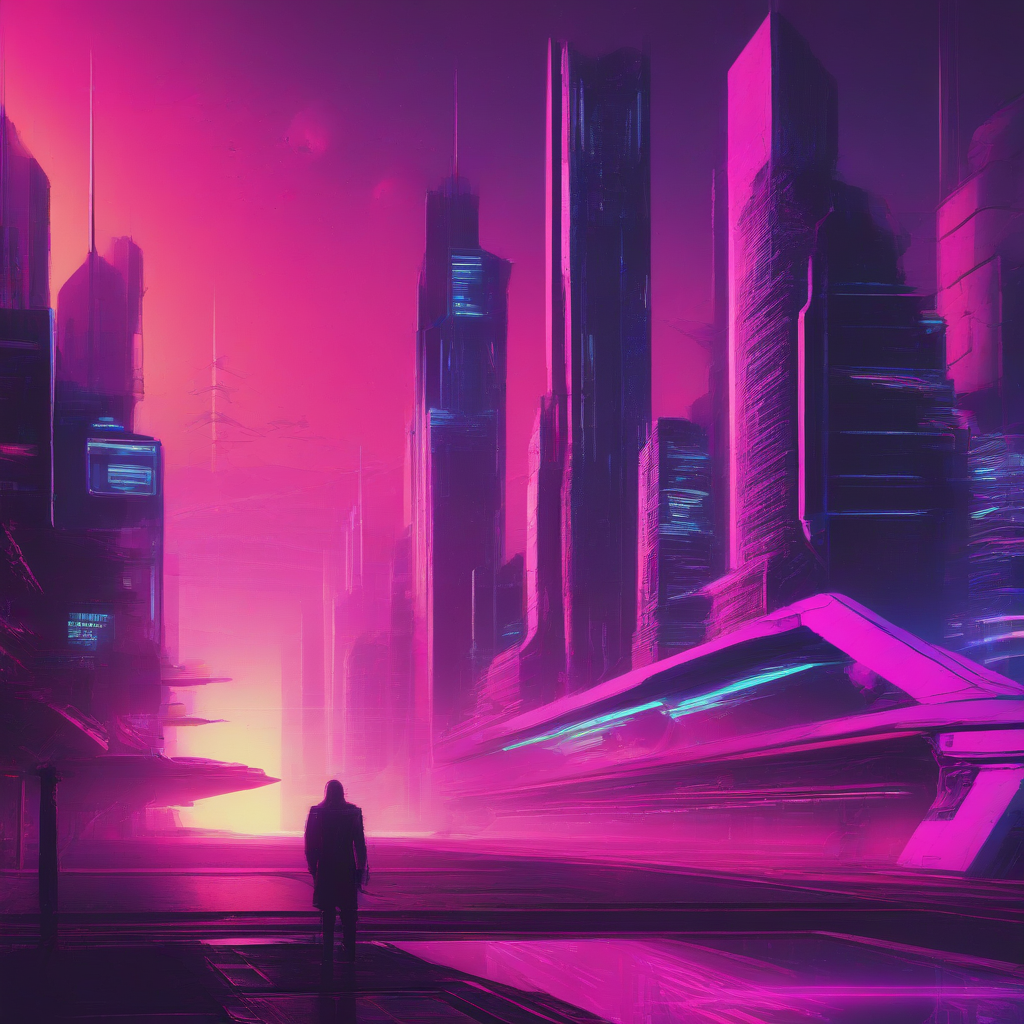} \\
\raisebox{0.95cm}{\makecell[tt]{\textbf{DDIM} \\  \textit{13-step}}} &  
\includegraphics[width=0.13\textwidth]{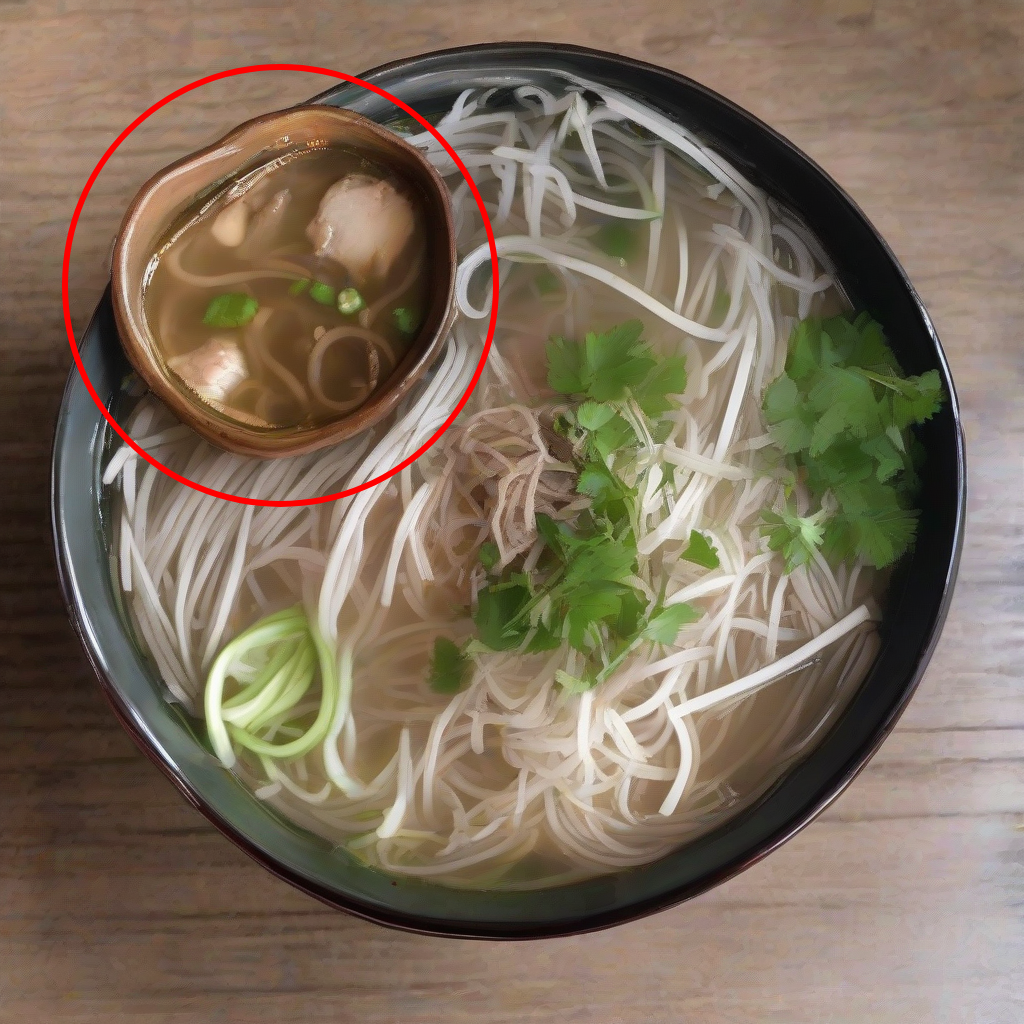} & \includegraphics[width=0.13\textwidth]{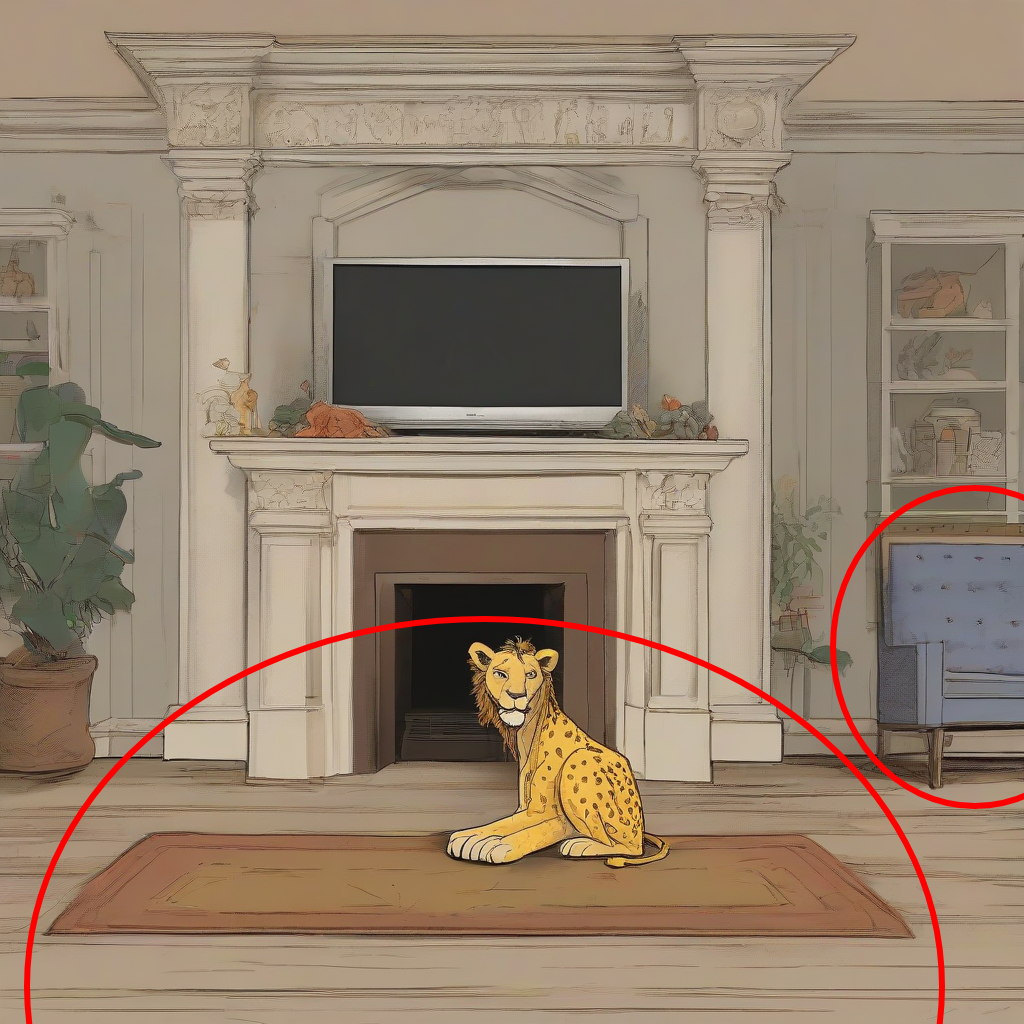} & \includegraphics[width=0.13\textwidth]{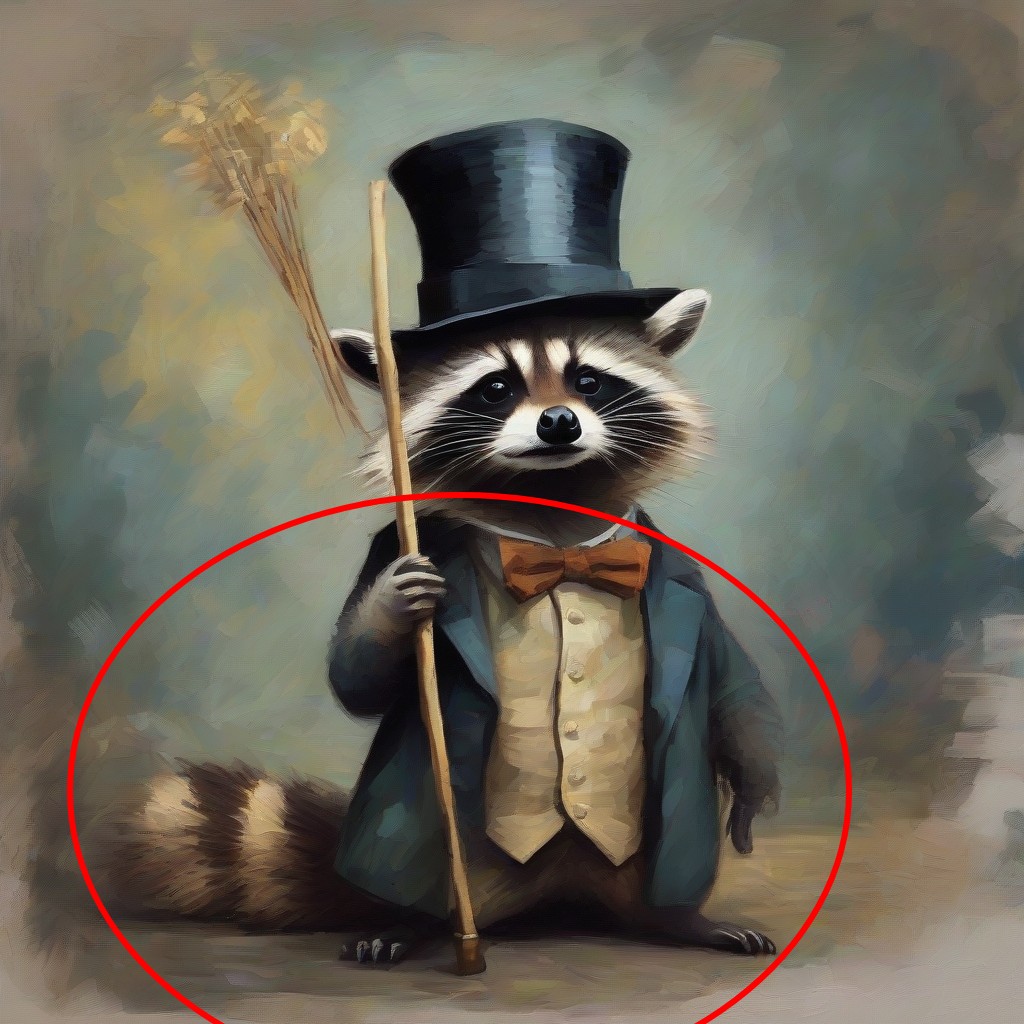} & \includegraphics[width=0.13\textwidth]{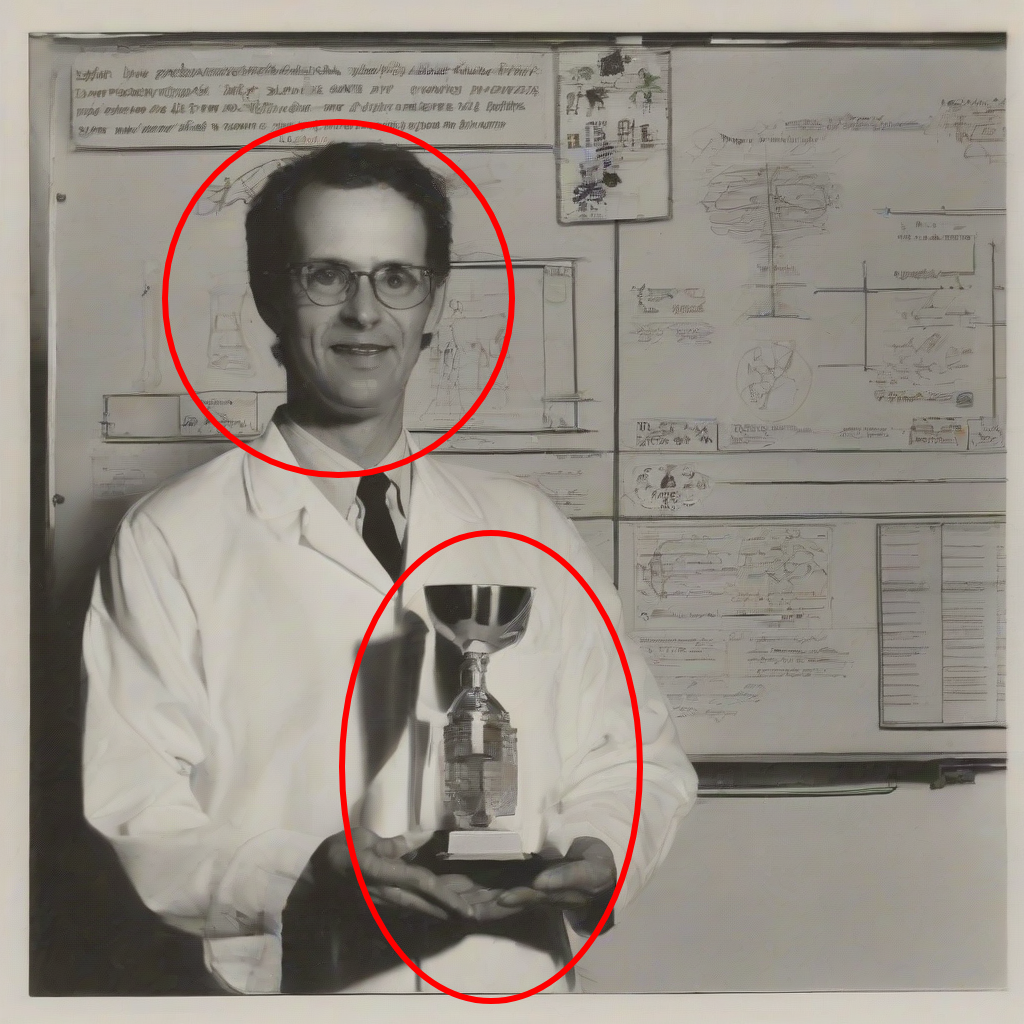} & \includegraphics[width=0.13\textwidth]{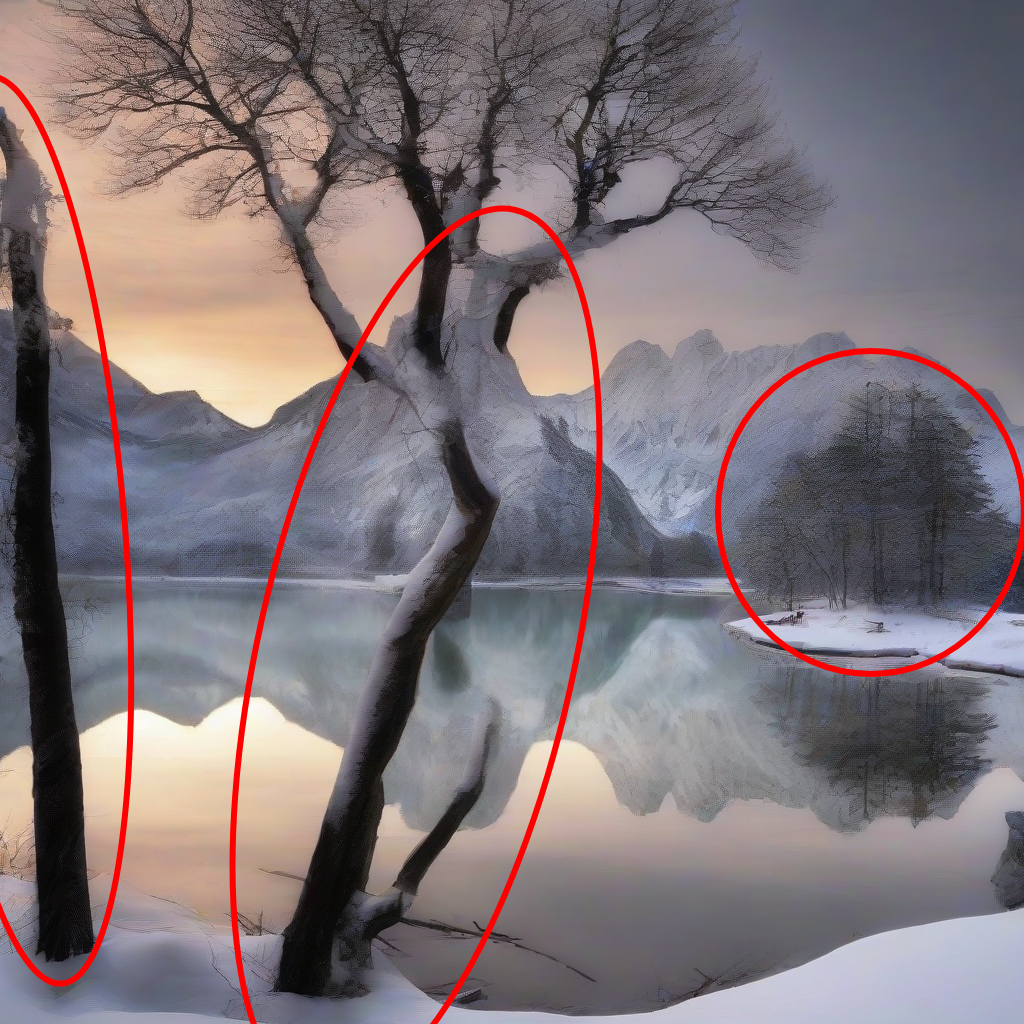} & \includegraphics[width=0.13\textwidth]{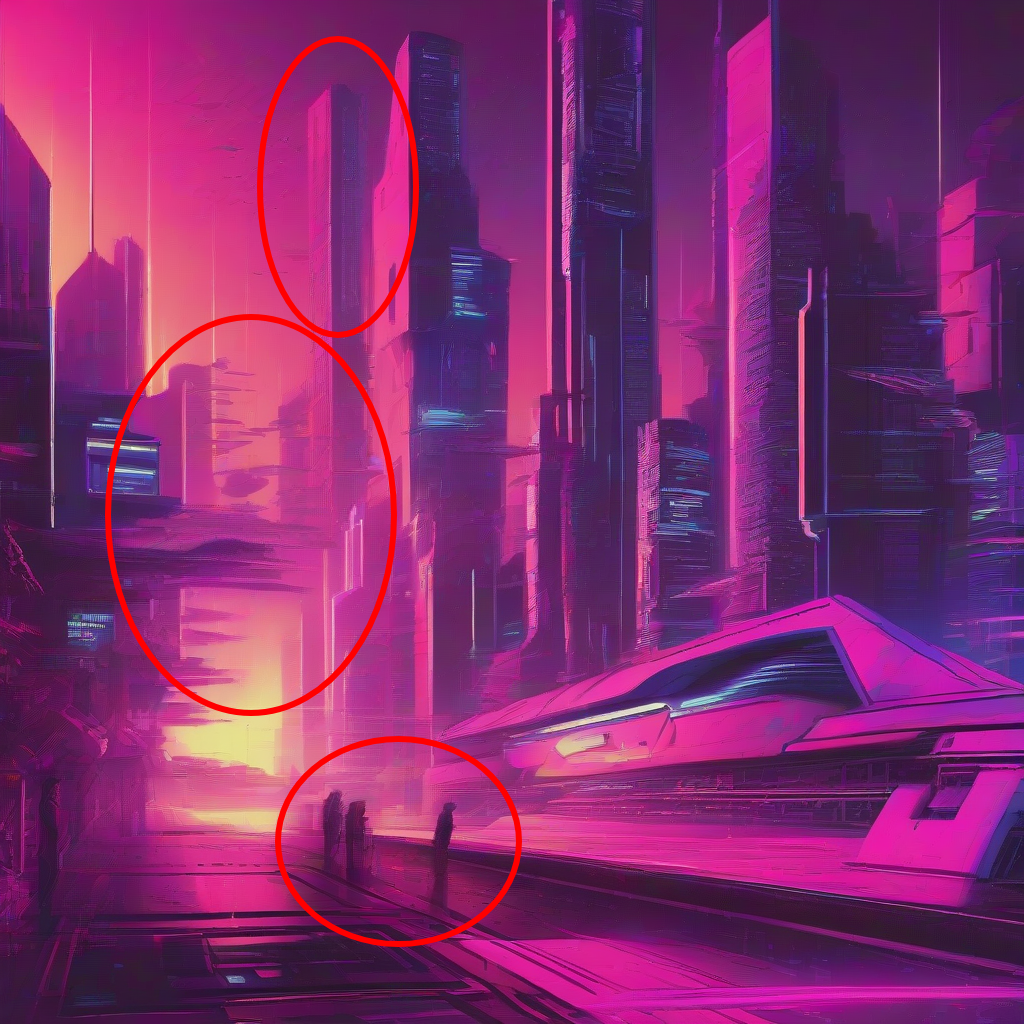} \\
\end{tabular}
\end{minipage}
\caption{This figure compares a step-reduced sampler with our best reuse strategy of (approximately) the same latency.
The reuse strategy clearly outperforms the step-reduced sampler at producing realistic images that match the original 20-step sampler. \added{Red circles have been added to the images on the bottom row to help readers identify some of the key differences between the 20-step and 13-step DDIM sampler.}}
\label{fig:1}
\end{figure*}

\section{Introduction}\label{sec1}
\noindent
Diffusion probabilistic models (DPMs) have become increasingly popular for text-conditioned image generation  \cite{ramesh2022hierarchical, dhariwal2021diffusion, nichol2021glide, saharia2022photorealistic}.
While DPMs can generate images of unprecedented quality, they require considerable amounts of time in order to do so, motivating researchers to improve their efficiency.
Currently, there are two main approaches for improving DPM efficiency: (1) decrease the number of calls to the U-Net, and (2) decrease the cost of calling the U-Net.\footnote{A U-Net is the deep neural network that powers DPMs.} 
This is typically facilitated by knowledge distillation or alterations to the U-Net's training objective.
Notably, there has been minimal work on methods to improve a DPM's latency without any retraining, fine-tuning, or knowledge distillation.

\vspace{0.5\baselineskip}
\noindent
Our paper focuses on this gap, by directly removing an expensive aspect of \deleted{the}\added{DPMs'} sampling procedure that we find to be redundant: the repeated calculation of attention maps.
Specifically, instead of recalculating attention maps from the key-query pairs at each step, the most recently calculated attention maps are stored in memory and can be reused during the sampling procedure. 
The main contribution of this paper is identifying and examining the reuse strategies that produce the smallest distortions to the original image. In particular: 

\noindent
\begin{enumerate}
    \item We locate a heuristic reuse strategy by analysing error propagation in the reverse diffusion process.
    \item We adapt this heuristic strategy to account for dependencies between different steps of the sampling procedure that the heuristic strategy neglects. 
    \item We show that our reuse strategies outperform step-reduced samplers of comparable latency.
\end{enumerate}

\section{Related Work}\label{sec2}
\noindent
\textbf{Sampling from a Diffusion Model.}
The forward diffusion process iteratively injects noise into an input image\deleted{ in order to transform it}\added{, gradually transforming the original image} into a standrad Gaussian. 
In the reverse direction, de-noising involves repeatedly invoking a decoder, which is typically trained to predict a score function, $\nabla_{\boldsymbol{x}} \text{log } q_t (\boldsymbol{x}_t)$. This is either learned directly, $\boldsymbol{s}_{\theta}\left(\boldsymbol{x}_{t}, t\right)$, or indirectly via error estimation, $\boldsymbol{\epsilon}_{\theta}\left(\boldsymbol{x}_{t}, t\right)$. 
Once the network is trained, the model must be paired with a sampling procedure that uses information from the decoder to reproduce the original image from Gaussian noise, potentially with the aid of a prompt. 
The reverse diffusion process is often modelled by the following ODE \cite{song2020score}:
\begin{gather}
    \frac{d \boldsymbol{x}_{t}}{dt} = f(t) \boldsymbol{x}_{t} - \frac{1}{2} g^{2}(t)\nabla_{\boldsymbol{x}} \text{log } q_t (\boldsymbol{x}_t) ,
\end{gather}
\noindent
where $f(t)$, $g(t)$ are determined by the noise schedule. 
Successfully solving this equation corresponds to the DPM sampling a realistic image.

\vspace{0.5\baselineskip}
\noindent
\textbf{Decreasing the Number of U-Net Calls.}
A DPM's latency is often reduced by lowering the total number of steps in its sampling procedure.
But by shortening a DPM's sampling process, a developer might degrade its performance, as shown in Figure \ref{fig:1}.
Therefore, researchers typically use knowledge distillation to maintain sample quality, training a new (step-reduced) network to emulate the original (high-latency) DPM.
For example, progressive distillation \cite{salimans2022progressive, meng2023distillation} iteratively trains a student network to predict the original DPM's two-step output in a single step, repeatedly halving the number calls to the U-Net.
Since then, a variety of new methods have been proposed to extend this simple approach.
Several researchers \cite{song2023consistency, gu2023boot, berthelot2023tract} have introduced a notion of consistency, producing single-step solvers that generate (or sample from) any point along the whole diffusion trajectory. 
Alternatively, Liu et al. \cite{liu2023instaflow} reduced the curvature of the score function, which made the U-Net more accurate for large step-sizes. 

\vspace{0.5\baselineskip}
\noindent
All of the methods listed above reduce latency with a minimal impact on sample quality.
However, they require some form of (re-)training and, as such, \deleted{can't}{cannot} directly bolster pre-trained DPMs.
In contrast, Lu et al. \cite{lu2022dpm} leveraged ODE-theory to directly improve DPM's sampling procedure. 
Specifically, they derived an analytical solution to the reverse diffusion ODE expressed as an exponentially weighted integral.
Based on this, they developed a numerical method that approximates the Taylor expansion of the exponential integral, allowing the U-Net to track the curvature of \added{the} score function with larger step-sizes \cite{dockhorn2022genie}.
Unlike the approaches that require knowledge distillation, this method can be directly applied to improve pre-trained DPMs in a `plug-and-play' manner \cite{lu2022dpm}.

\vspace{0.5\baselineskip}
\noindent
\textbf{Decreasing the Cost of U-Net Calls.}
Some researchers focus on reducing the cost of a single step rather than reducing the total number of steps.
In particular, they perform knowledge distillation by training a small (i.e. low-cost) U-Net to emulate the large U-Net used by the original DPM. 
For example, Kim et al. \cite{kim2023architectural} manually removed blocks from the U-Net that powers Stable Diffusion \cite{rombach2022high} and trained \deleted{it}\added{this pruned network} to approximate \deleted{the unpruned network}\added{its unpruned counterpart}.
Li et al. \cite{li2023snapfusion} developed this manual approach by pruning the U-Net in accordance with a formalised trade-off between performance (CLIP-score) and latency.
Both of these papers retain sample quality by employing knowledge distillation, but can the cost of calling a U-Net be reduced without (re-)training?

\vspace{0.5\baselineskip}
\noindent
Bhojanapalli et al. \cite{bhojanapalli2021leveraging} improved transformers' efficiency by exploiting redundancies in their repeated calculation of attention maps. 
These maps are particularly consequential for latency as their calculation involves a costly outer product between a high-dimensional key and query.
Upon finding a reasonable degree of similarity between a transformer's attention maps, they reused an arbitrary subset of the first layer's maps in the following layers.
This approach can be applied in a `plug-and-play' manner to any pre-trained transformer, as it \deleted{doesn't}\added{does not} require any retraining or knowledge distillation. 
We wonder whether similar redundancies exist in the attention maps produced \deleted{by DPMs}\added{during text-to-image diffusion}, and if so, whether this can be exploited to reduce the cost of calling a DPM's U-Net.

\section{Analysis}\label{sec3}
We start this section by examining redundancies in the repeated calculation of \added{DPM's} attention maps \deleted{by DPMs}.
Upon finding a significant degree of redundancy, we then define and locate reuse strategies that attempt to optimally exploit this redundancy. 
In particular, we propose a simple \added{reuse} strategy by considering the Lyapunov exponents (a common tool for studying ODEs) of the reverse diffusion process. 
All of the figures presented in this section are run using Stable Diffusion v1.5 with 20-step DDIM \cite{song2020denoising} as the \deleted{(base) sampling procedure}\added{sampler}.

\begin{figure*}[htbp]
    \centering
        \begin{minipage}[b]{\textwidth}
         \centering\includegraphics[width=0.49\textwidth]{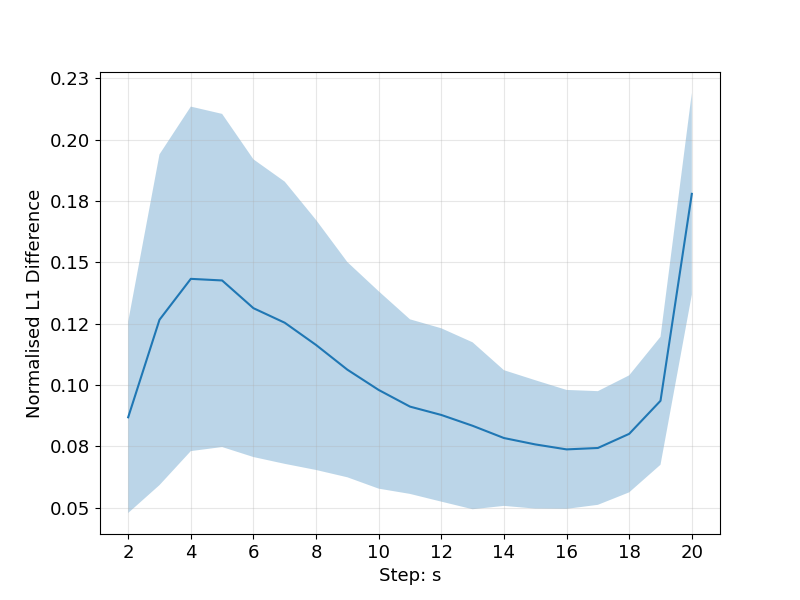}
         \centering\includegraphics[width=0.49\textwidth]{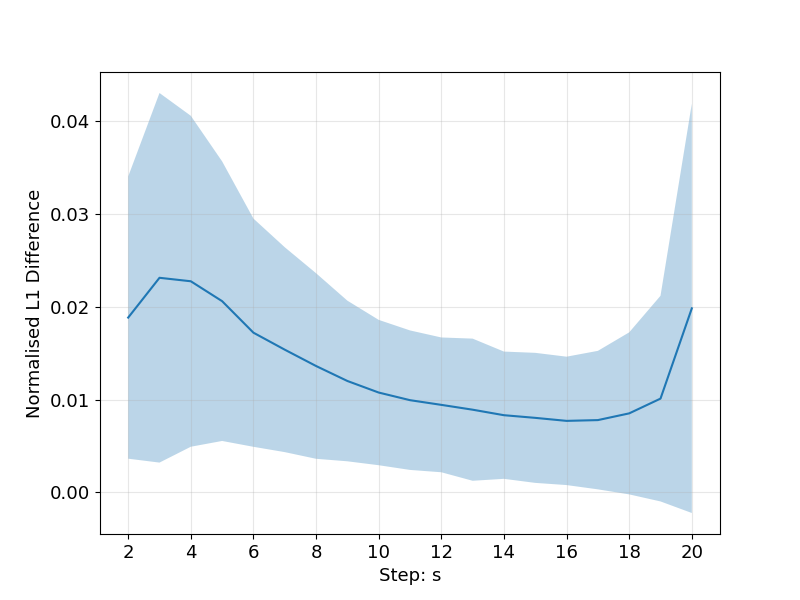}
    \end{minipage}
    \caption{Normalised L1-distance between the Self- (left) and Cross- (right) attention maps $A(s)$ and $A(s-1)$ for an unperturbed flow. This is generated from 200 random ImageNet prompts. The shaded region includes one standard deviation.}
    \label{fig:Similarity}
\end{figure*}

\vspace{0.5\baselineskip}
\noindent
\textbf{Attention Map Redundancy.}
We conjecture that DPM attention maps are (predictably) similar to their temporally adjacent neighbours. 
As such, we empirically investigate the normalised L1 difference \cite{bhojanapalli2021leveraging} between them, averaged over a variety of prompts. 
Figure \ref{fig:Similarity} illustrates a high degree of similarity, in which temporally adjacent maps are (on average) within 0.2 of each other.
This suggests a relatively stable degree of redundancy in the repeated calculation of attention maps that can be exploited to reduce 
a DPM's latency.
Instead of repeatedly calculating attention maps from key-query pairs, certain maps can be used more than once. 
But what exactly would this look like?

\vspace{0.5\baselineskip}
\noindent
\textbf{Defining Reuse and Success.} 
Let $\boldsymbol{A}_s^l$ denote the attention map calculated in the $s$-th call of the $l$-th layer of the U-Net in the reverse diffusion process.\footnote{\deleted{In fact, the}\added{The} conditioned and unconditioned prompts both have a corresponding cross- and self- attention map. For brevity, we refer to just one map at each layer and step.} 
Moreover, let $\boldsymbol{M}^l$ denote a set of memory variables which store the maps that we may wish to reuse. 
During sampling, set $\boldsymbol{M}^l \leftarrow \boldsymbol{A}_s^l$ every time that an attention map is directly calculated from key-query pairs. 
Then, for a reuse step $r>1$, set $\boldsymbol{A}_r^l \leftarrow \boldsymbol{M}^l$, instead of directly calculating $\boldsymbol{A}_r^l$ from key-query pairs. 
We parameterise a reuse strategy by a binary vector, $\boldsymbol{\pi}$, where each entry corresponds to a step in the sampling procedure. 

\vspace{0.5\baselineskip}
\noindent
In particular, $\boldsymbol{\pi}_s=\texttt{1}$ indicates that the attention map $\boldsymbol{A}_s$ is directly calculated from a key-query pair and $\boldsymbol{\pi}_s=\texttt{0}$ denotes reuse at sampling step $s$.
So\added{,} the reuse strategy $\boldsymbol{\pi}=\left[\texttt{1, 1, 0, 0, 1, 0}\right]$ for a 6-step sampling procedure would directly calculate the attention maps from key-query pairs at step\added{s} 1, 2 and 5. 
Consequently, it would reuse the second attention map in steps 3 and 4 and reuse the fifth attention map at step 6.

\vspace{0.5\baselineskip}
\noindent
While we have now defined what reuse involves, it is also important to consider when it can be deemed successful.
In this paper, we consider a reuse strategy to be successful if it generates images that are close to the original DPM's output given the same prompt and initial conditions, as measured by PSNR.
A looser definition of success would require only that reuse strategies generate realistic outputs that align with their prompt.
This could be evaluated by measuring the outputs' CLIP-Score or FID with a distribution of natural images. 

\vspace{0.5\baselineskip}
\noindent
The PSNR-measurable definition of success is preferable as \deleted{it's}\added{it is} easier to evaluate and less subjective. 
In particular, the PSNR between two samples can be calculated quickly, which facilitates fast search algorithms. 
Moreover, PSNR \deleted{doesn't}\added{does not} depend on an arbitrary choice of natural images, unlike FID and CLIP.
For the reasons outlined above, our reuse strategies are optimised (via PSNR) to approximate the behaviour of a normal DPM, which itself could be trained to optimise FID or CLIP-Scores. 
Nevertheless, in Section \ref{sec5} we present results regarding both characterisations of success.

\vspace{0.5\baselineskip}
\noindent
\textbf{Locating a Successful Reuse strategy.} 
\added{Given a fixed sampling procedure and desired latency,} a brute force approach for finding \deleted{a successful}\added{the best} reuse strategy would \deleted{evaluate}\added{test} all \deleted{possibilities}\added{possible strategies (i.e., all possible binary vectors, $\boldsymbol{\pi}$) at or below the desired latency} and select the one whose output is closest to the true sample.
However, an exhaustive search is infeasible given the exponential growth in possible strategies as the number of steps increases. 
As such, in this section, we probe the diffusion process for insights that might warm-up our search. 
\added{To this end, we consider} a common tool for analysing dynamical systems ($\textbf{z}_t$)\deleted{ are}\added{, namely} `Lyapunov exponents' \cite{layek2015introduction} \added{---} which assume that errors ($\delta \textbf{z}_{t}$) grow (or shrink) exponentially in time. 
Formally, they assume there exists a $k\in \mathbb{R} \backslash \{0\}$ such that for every $t>\tau>0$:
\begin{equation}
    \left| \delta \textbf{z}_t\right| \approx e^{k \tau} \left| \delta \textbf{z}_{t-\tau} \right|
\end{equation}
For our case, let $\boldsymbol{x}_t$ represent the reverse diffusion process, where $\boldsymbol{x}_0$ is Gaussian noise and $\boldsymbol{x}_T$ is a sample. 
Furthermore, let $\textbf{A}_{s}$ denote the attention maps at sampling step s.
Adapting Lyapunov exponents for a reverse diffusion process in which attention maps are perturbed, we conjecture that there exists a $\lambda>0$ such that for every step, $s\in \mathbb{N}$ and corresponding timestep $t_s$ with $T>t_s>0$:
\begin{equation}
\label{eq}
    \left| \delta \boldsymbol{x}_T\right| \propto e^{\lambda (T-t_s)} \left| \delta \textbf{A}_{s}\right| \propto e^{-\lambda t_s} \left| \delta \textbf{A}_{s}\right|
\end{equation}
That is, the change in the final sample ($\delta \boldsymbol{x}_T$) is proportional to the change in the attention maps ($\delta \textbf{A}_s$) introduced by reuse at step $s$, subject to an exponential growth over the remaining steps.
If our conjecture is true, then a heuristic strategy that follows from this would be to reuse (i.e., perturb the attention map) as late as possible in the sampling procedure, maximising $s$. We will refer to this \textbf{H}e\textbf{UR}istic \textbf{R}euse Strateg\textbf{Y} as HURRY, defined for $r$ reuse steps in an a N-step sampler by:
\begin{equation}
    \text{HURRY} =  \left[\textbf{1} | \textbf{0} \right] = \Bigl[ \overbrace{\texttt{1, \ldots,  1}}^{N-r}, \overbrace{\texttt{0, \ldots,  0}}^{r} \Bigr]
\end{equation}

\begin{figure*}[htbp]
    \centering
    \begin{minipage}[b]{\textwidth}
        \centering
        \includegraphics[width=\textwidth]{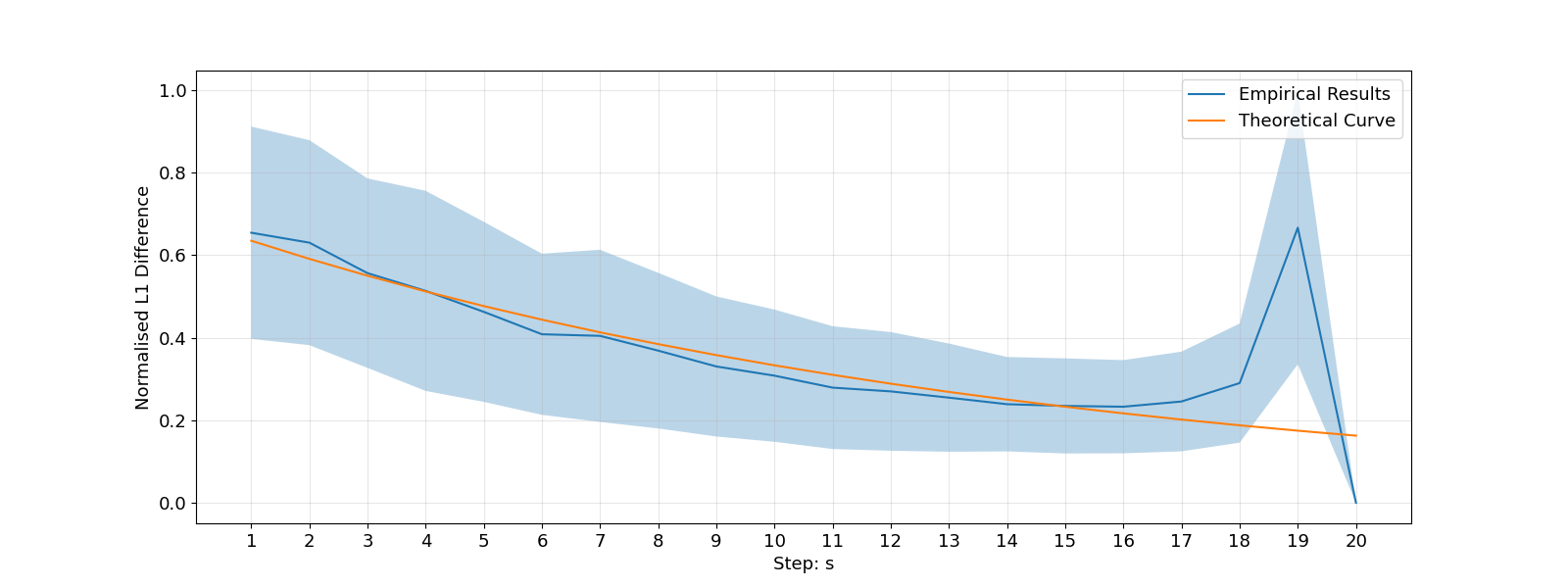}
    \end{minipage}
    \caption{The L1-distance between a sample produced by a normal DPM and a DPM where the attention map is perturbed at sampling step s. Specifically, the pre-softmax attention map is perturbed (in proportion to its norm) at step s. The results are scaled into the range [0,1] and then averaged over 200 random ImageNet prompts; the shaded region includes one standard deviation. The orange curve is of the form $k_1e^{-k_2s}$, tuned to approximate the empirical results between steps 1 and 18 (inclusive). }
    \label{fig:Pert}
\end{figure*}

\vspace{0.5\baselineskip}
\noindent
Equally, if $\lambda$ were negative, then the error would shrink exponentially, suggesting that reuse is most suitable early in the sampling procedure. 
Figure \ref{fig:Pert} empirically explores whether the (attention-adapted) Lyapunov exponent appears to be positive or negative.
We find that the impact of perturbations to the U-Net's attention maps decrease (roughly) monotonically over the sampling procedure, at least between steps one and eighteen (inclusive).
The correlation between the empirical data and the theoretical curve (exponential decay) is 0.96, for the first eighteen steps.
This decay supports our conjecture that the (attention-adapted) Lyapunov exponent of the reverse diffusion process is positive, see Equation \ref{eq}.

\section{Method}\label{sec4}
In \deleted{the} Section \ref{sec3}\added{,} we developed a reuse strategy based on the following conjectures: (1) The temporally-adjacent attention maps of a pre-trained DPM are (predictably) similar to each other; (2) The (attention-adapted) Lyapunov exponent of a pre-trained DPM is positive.
The first conjecture implies that reuse \added{is} a sensible idea\added{,} and the second implies that if you are going to reuse, it is best to do it late into the sampling procedure.
We provided supporting evidence for each of these conjectures in Figures \ref{fig:Similarity} and \ref{fig:Pert}\deleted{, respectively}.

\vspace{0.5\baselineskip}
\noindent
However, HURRY also relies on an assumption that we have so far neglected.
In particular, it assumes that the suitability of step $s$ for reuse is independent of whether the model has already reused at step $r<s$.
This might be problematic when creating reuse strategies that contain more than one reuse step.
For instance, a `later is better' strategy might be ideal for a single instance of reuse.
But, clustering several reuse steps towards the end of the sampling procedure without any intermediate recalculations \added{of attention maps} might be sub-optimal.

\vspace{0.5\baselineskip}
\noindent
\textbf{Perturbing HURRY.}
To address this limitation, we evaluate perturbations of our heuristic reuse strategy to determine whether they perform unexpectedly well due to some unforeseen dependencies. 
Specifically, we set HURRY as our initial `Best Strategy' and then perform a greedy search to locate the closest locally optimal strategy. 
A reuse strategy is said to be locally optimal if it outperforms all strategies that are one bit-flip away; where a bit-flip swaps a reuse and non-reuse step.\footnote{This is the smallest strategy mutation that preserves the number of reuse steps.}
We refer to the result of this search algorithm as PHAST: \textbf{P}erturbed \textbf{H}euristic \textbf{A}ttention \textbf{ST}rategy - this is our second reuse strategy.
The utility function below, $U$, is PSNR in our case. 

\begin{minipage}[b]{\textwidth}
\centering
\begin{minipage}[b]{0.7\textwidth}
\begin{algorithm}[H]
\caption{Search Algorithm for PHAST}
\label{alg:PHAST}
$\text{Best Strategy} \gets HURRY$
\newline $Optima(0) \gets U(\text{Best Strategy})$ 
\newline $Run \gets 0$
\newline\textbf{repeat}
\newline \textcolor{white}{-} \quad $Run \gets Run + 1$
\newline \textcolor{white}{-} \quad $Optima(Run) \gets Optima(Run-1)$
\newline \textcolor{white}{-} \quad \textbf{for} $\boldsymbol{\pi} \in BitFlipSet(\text{Best Strategy})$
\newline \textcolor{white}{-} \quad \textcolor{white}{-} \quad \textbf{if} $U(\boldsymbol{\pi})>Optima(Run)+\varepsilon$
\newline \textcolor{white}{-} \quad \textcolor{white}{-} \quad \textcolor{white}{-} \quad $\text{Best Strategy} \gets \boldsymbol{\pi}$
\newline \textcolor{white}{-} \quad \textcolor{white}{-} \quad \textcolor{white}{-} \quad $Optima(Run) \gets U(\boldsymbol{\pi})$
\newline \textcolor{white}{-} \quad \textcolor{white}{-} \quad \textbf{end if}
\newline \textcolor{white}{-} \quad \textbf{end for}
\newline \textbf{until} $Optima(Run) = Optima(Run-1)$
\newline \textbf{return}  $\text{Best Strategy}$
\end{algorithm}
\end{minipage}
\end{minipage}

\vspace{0.5\baselineskip}
\noindent
Algorithm \ref{alg:PHAST} has a number of desirable properties. 
One such property is that, unlike evolutionary approaches, \deleted{it's}\added{it is} deterministic.
This determinism stems from the fact that it (effectively) performs gradient descent on the set of reuse strategies, calculating the finite difference between adjacent strategies w.r.t. to PSNR, and moving to (approximately) minimise this.\footnote{We include a small threshold, $\varepsilon$, to prevent insignificant rounds of bit flipping.}  
Another desirable property is that \deleted{it's}\added{it is} quick. 
For an N-step sampling procedure with r reuse steps, there are \deleted{only} $\mathcal{O}(N^2)$ bit-flipped strategies, in comparison to an exhaustive search, which has $\mathcal{O}(N^r)$ strategies.
In the case of a 20-step sampling procedure with 10 reuse steps, this corresponds to one round of bit-flipping covering 100 strategies, while an exhaustive search must cover 184,756 strategies. 

\section{Evaluation}\label{sec5}
In this section, we begin by examining the effectiveness of algorithm \ref{alg:PHAST} at locating the optimal reuse strategy.
Following this, we compare the performance of attention reuse and step-reduction \added{(of comparable latency)} across various datasets, models, and sampling procedures.
Our results demonstrate that reuse outperforms step-reduction in terms of PSNR and is competitive w.r.t. FID and CLIP-Score.

\vspace{0.5\baselineskip}
\noindent
\textbf{Comparison with Alternative Reuse Strategies.}
How does algorithm \ref{alg:PHAST} compare to an exhaustive search?
To investigate this, we evaluated the PSNR of every reuse strategy (averaged over ten prompts from ImageNet) in the space of 10-step DPM++ samplers with 3 reuse steps.
As a result of this exhaustive search, we found HURRY to be the third-best strategy, with the two best strategies differing by a single bit-flip:
\begin{enumerate}
\item \text{Strategy = \texttt{[1,1,1,1,1, 1,0,1,0,0]}, PSNR = 27.2} (i.e., PHAST)
\item \text{Strategy = \texttt{[1,1,1,1,1, 1,0,0,1,0]}, PSNR = 26.8}
\item \text{Strategy = \texttt{[1,1,1,1,1, 1,1,0,0,0]}, PSNR = 25.9} (i.e., HURRY)
\end{enumerate}
While we \deleted{can't}\added{cannot} perform an exhaustive search on larger spaces\added{,} we find that algorithm \ref{alg:PHAST} rapidly settles into a local optima, typically within a single bit-flip.
For instance, given a 20-step DDIM sampler with 10 reuse steps, algorithm \ref{alg:PHAST} terminates \added{at:}\\
\\
PHAST =  \texttt{[1,1,1,1,1, 1,1,1,0,1, 0,0,0,1,0, 0,0,0,0,0]}
\\

\begin{figure*}[h]
    \centering
      \centering
      \setlength{\tabcolsep}{3pt}
      \begin{tabular}{c c c c c c}
      \toprule
        \textbf{Strategy} & \textcolor{white}{-}\textbf{No Reuse} & \textbf{PHAST} & \textbf{HURRY} & \textbf{Random} & \textbf{Heuristic 1} \\
        PSNR ($\uparrow$) & \textcolor{white}{--}ref. & 27.10 & 24.56 & 19.88 & 21.10\\
        \raisebox{1.1cm}{\makecell[tt]{\textcolor{white}{-}Sample:\textcolor{white}{--}\\ \textit{Sealyham}\textcolor{white}{-}}} 
           & \textcolor{white}{-}\includegraphics[width=0.145\textwidth]{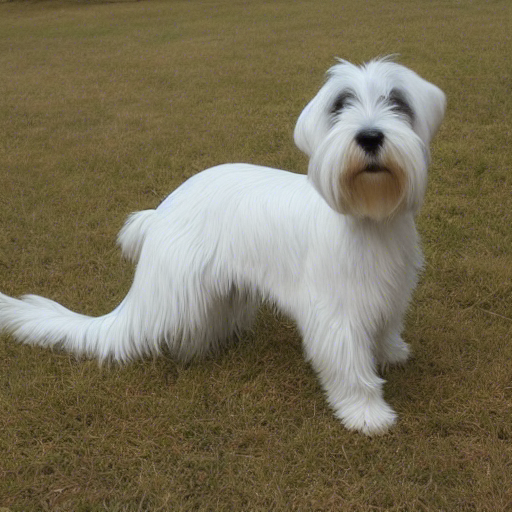} 
           & \includegraphics[width=0.145\textwidth]{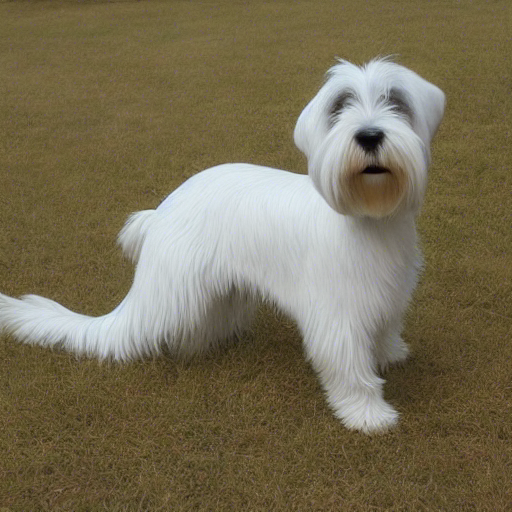}
           & \includegraphics[width=0.145\textwidth]{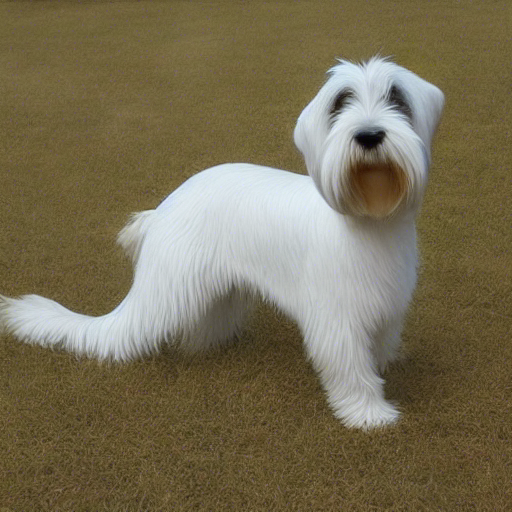}
           & \includegraphics[width=0.145\textwidth]{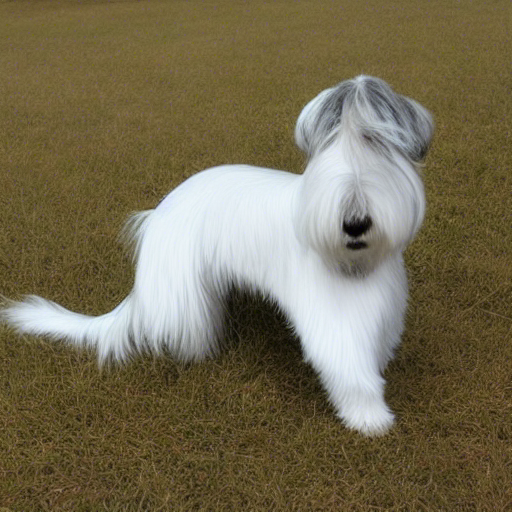}
           & \includegraphics[width=0.145\textwidth]{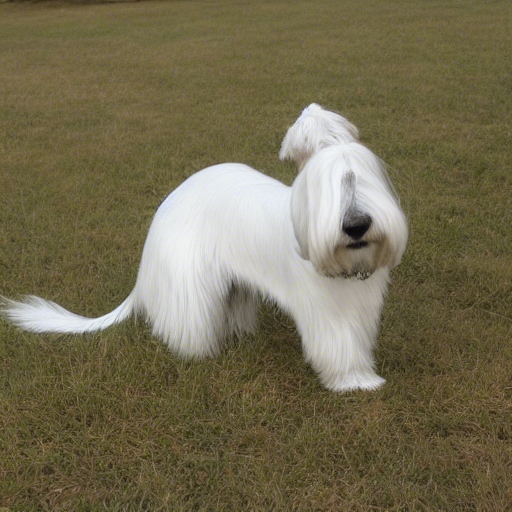} \\
      \end{tabular}

      \centering
      \setlength{\tabcolsep}{3pt}
      \begin{tabular}{c c c c c c}
      \midrule
        \textbf{Strategy} & \textcolor{white}{-}\textbf{Heuristic 2} &  \textbf{Heuristic 3} & \textbf{Heuristic 4} & \textbf{Heuristic 5} & \textbf{Heuristic 6} \\
        PSNR ($\uparrow$) & \textcolor{white}{--}16.78 & 18.07 & 21.18 & 18.03 & 19.61\\
        \raisebox{1.1cm}{\makecell[tt]{\textcolor{white}{-}Sample:\textcolor{white}{--}\\ \textit{Sealyham}\textcolor{white}{-}}} 
           & \textcolor{white}{-}\includegraphics[width=0.145\textwidth]{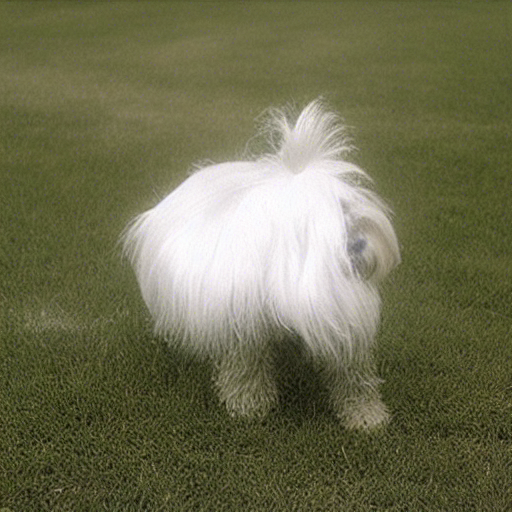} 
           & \includegraphics[width=0.145\textwidth]{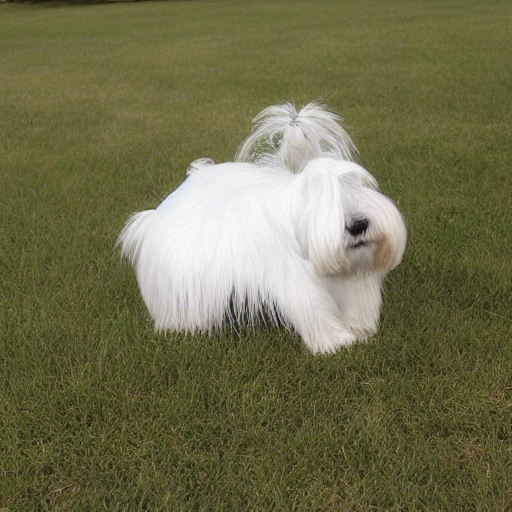}
           & \includegraphics[width=0.145\textwidth]{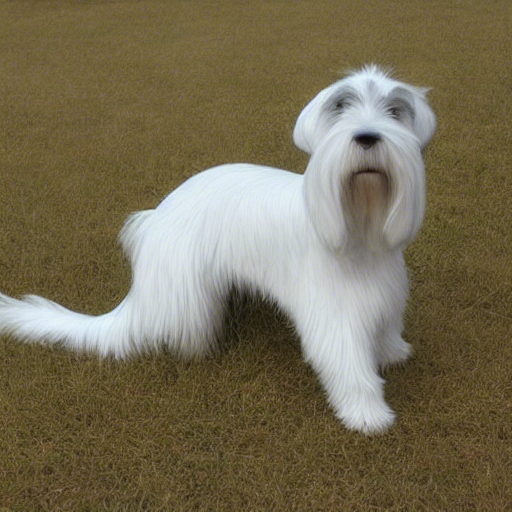}
           & \includegraphics[width=0.145\textwidth]{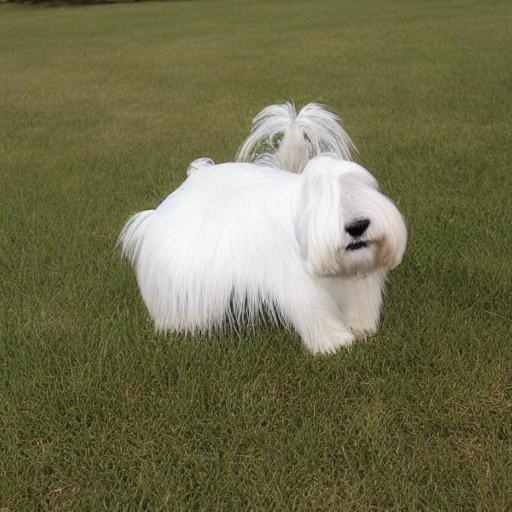}
           & \includegraphics[width=0.145\textwidth]{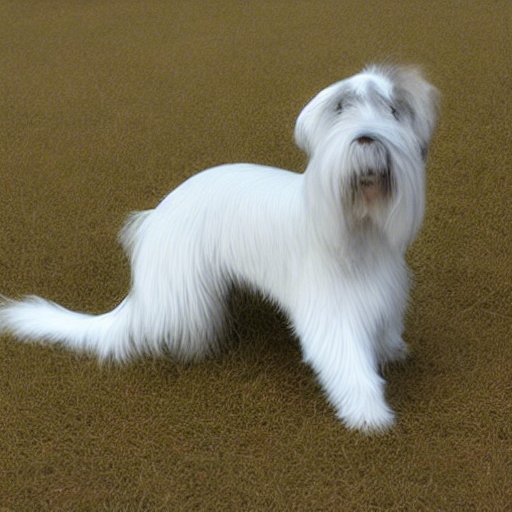} \\ \bottomrule 
      \end{tabular}

    \caption{\added{A comparison of PSNR (averaged over 200 ImageNet labels as prompts) between our reuse strategies and several alternative heuristic reuse strategies, for a 20-step DDIM sampler with 10 reuse steps. For qualitative analysis, a sample for the prompt `Sealyham terrier Sealyham' is provided for each strategy.}}
    \label{fig: Different}
\end{figure*}

\noindent
These \added{search} results indicate that HURRY is a near-optimal strategy, stunted by it\deleted{'}s assumption of step-wise independence, which PHAST amends. 
To further support this assessment, and given the infeasibility of an exhaustive search in larger spaces, we compare \deleted{our reuse strategies with several alternatives}\added{PHAST and HURRY with a small sample of heuristic (and randomised) reuse strategies} in Figure \ref{fig: Different}. \added{Notably, each of these alternative strategies performs worse than our reuse strategies, as they achieve a lower PSNR and are of a noticeably inferior quality to the human eye.}\footnote{\added{The heuristic reuse strategies in Figure \ref{fig: Different} each uses a simple alternating binary pattern, where the first entry is set to 1, as reuse cannot occur in the first step. Formally, they can be expressed as follows:\\(H1) \texttt{[1,0,0,1,0, 1,0,1,0,1, 0,1,0,1,0, 1,0,1,0,1]},
(H2) \texttt{[1,0,0,0,0, 0,0,0,0,0, 0,1,1,1,1, 1,1,1,1,1]}, \\ (H3) \texttt{[1,0,0,0,0, 0,1,1,1,1, 0,0,0,0,0, 1,1,1,1.1]}, (H4) \texttt{[1,1,1,1,1, 0,0,0,0,0, 1,1,1,1,1, 0,0,0,0,0]}, \\ (H5) \texttt{[1,0,0,0,0, 0,1,1,1,1, 1,1,1,1,1, 0,0,0,0,0]}, (H6) \texttt{[1,1,1,1,1, 0,0,0,0,0, 0,0,0,0,0, 1,1,1,1,1]}.}}

\begin{figure}[h]
    \centering
    \begin{minipage}[b]{\textwidth}
        \centering
        \includegraphics[width=\textwidth]{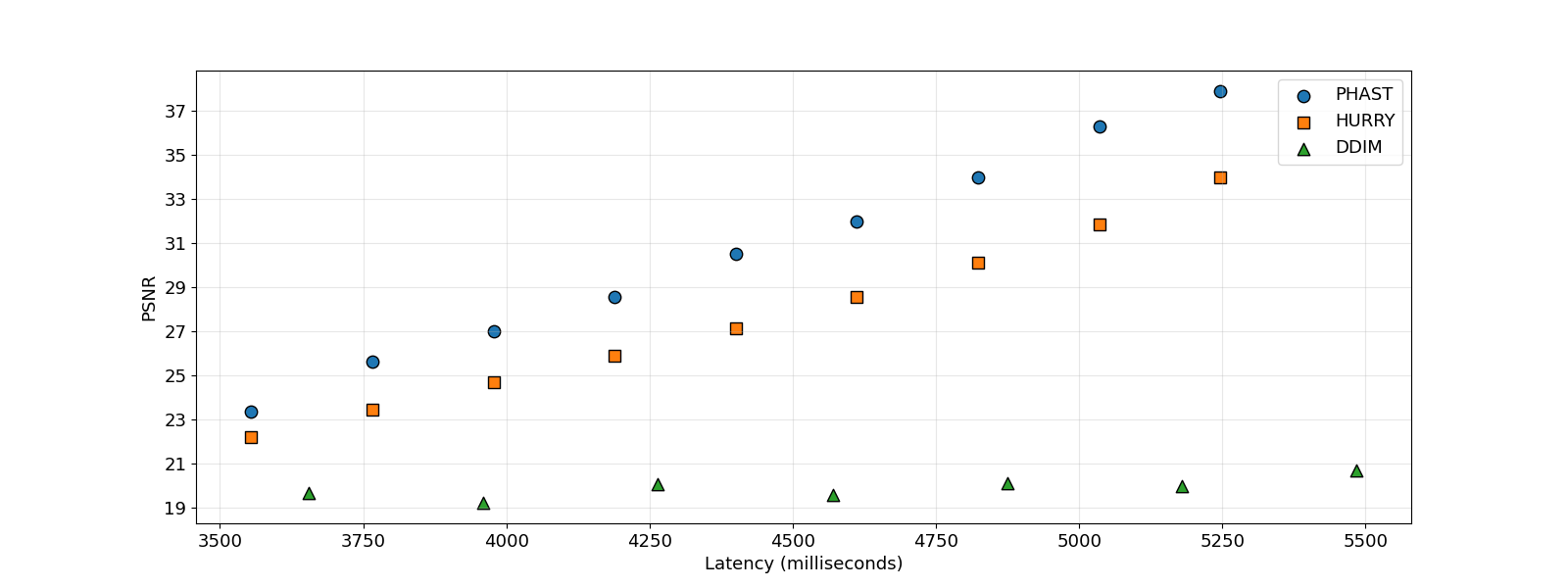}
    \end{minipage}
    \caption{This figure compares \deleted{the} PSNR \added{(averaged over 200 ImageNet labels as prompts)} \deleted{of}\added{for} PHAST, HURRY, and \added{the (base)} DDIM \added{sampler} for comparable latencies. \deleted{The PSNR is taken over 200 ImageNet labels as prompts.}}
    \label{fig:Results}
\end{figure}

\begin{figure*}[h]
  \centering
  \begin{minipage}{0.98\textwidth}
  \setlength{\tabcolsep}{3pt}
    \begin{tabular}{c c c c c c c}
      \raisebox{0.93cm}{\makecell[tt]{\textbf{DDIM} \\ \textit{20-step}}} & \includegraphics[width=0.13\textwidth]{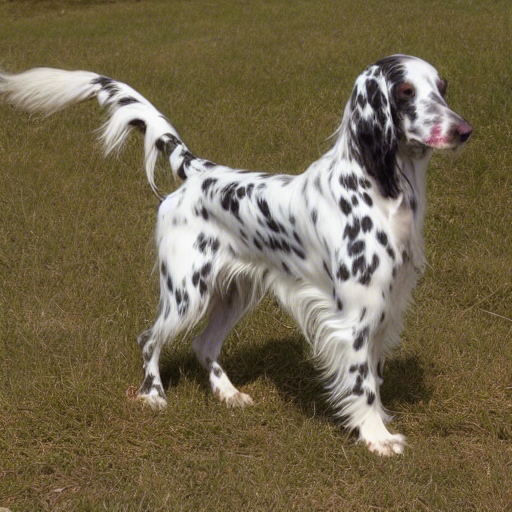} &
      \includegraphics[width=0.13\textwidth]{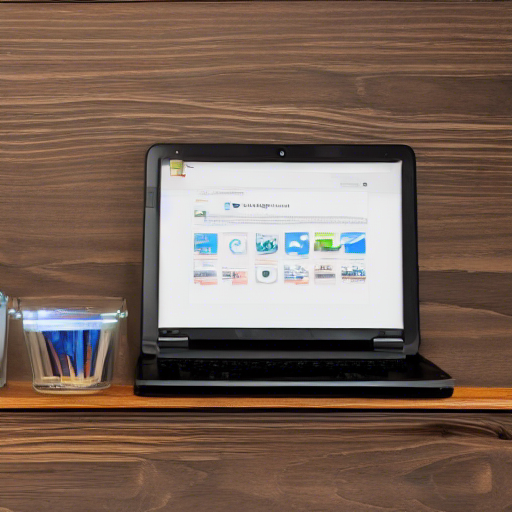} &
      \includegraphics[width=0.13\textwidth]{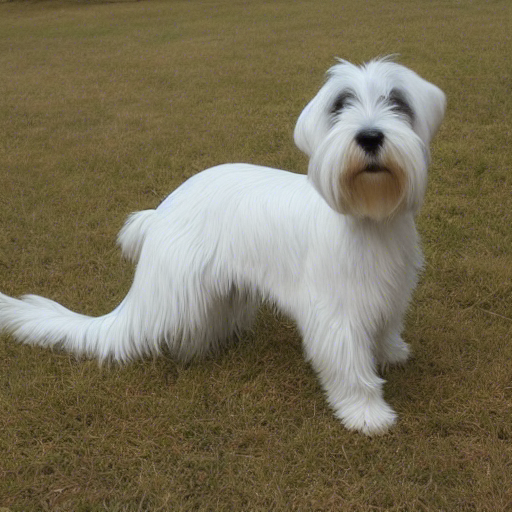} &
      \includegraphics[width=0.13\textwidth]{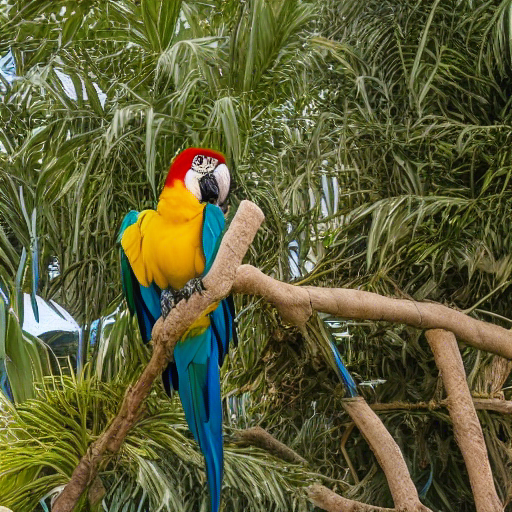} &
      \includegraphics[width=0.13\textwidth]{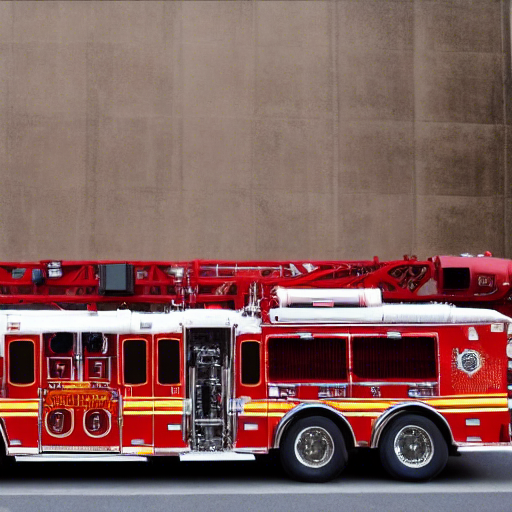} &
      \includegraphics[width=0.13\textwidth]{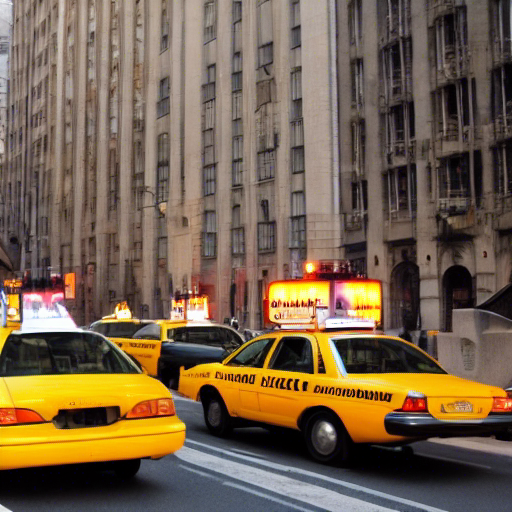} \\
      \raisebox{0.93cm}{\makecell[tt]{\textbf{DDIM} \\ \textit{12-step}}} & 
      \includegraphics[width=0.13\textwidth]{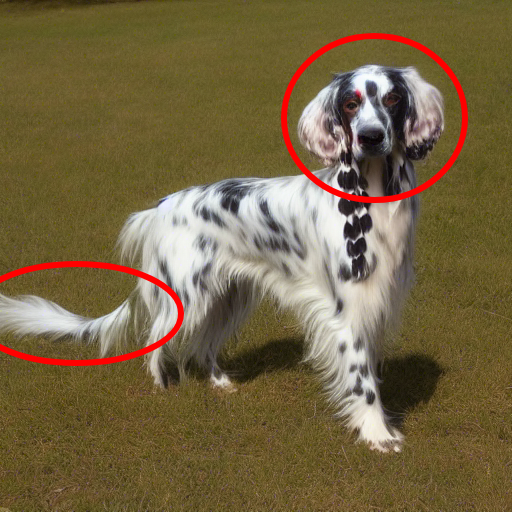} &
      \includegraphics[width=0.13\textwidth]{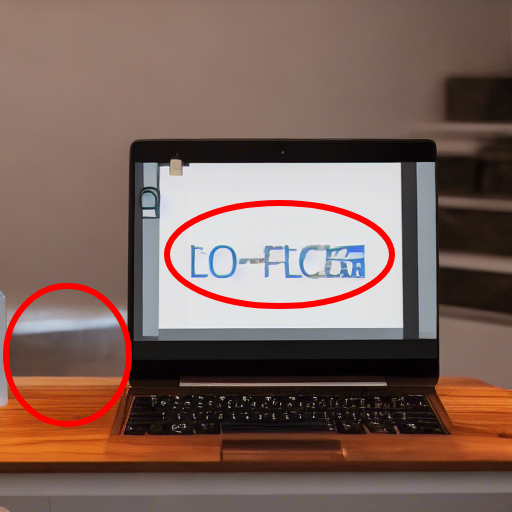} &
      \includegraphics[width=0.13\textwidth]{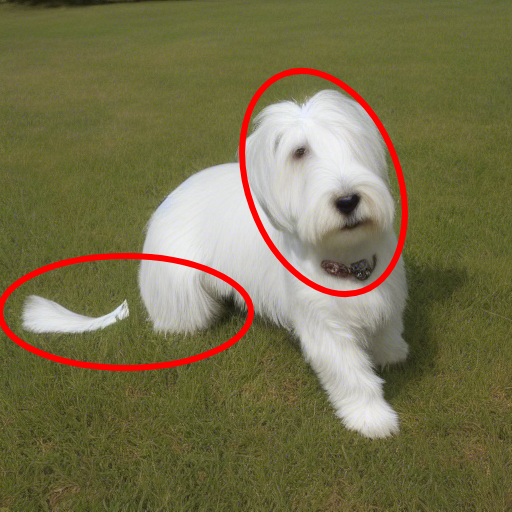} &
      \includegraphics[width=0.13\textwidth]{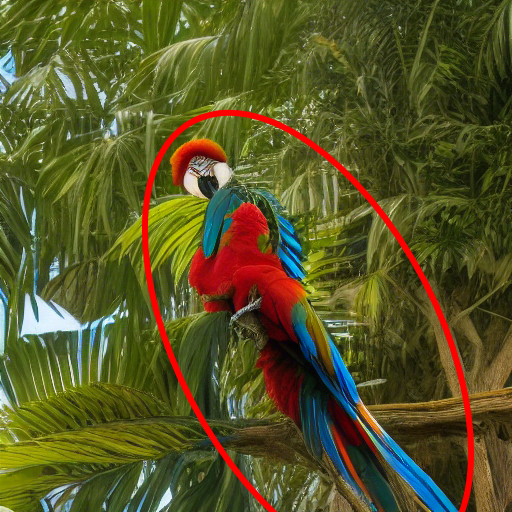} &
      \includegraphics[width=0.13\textwidth]{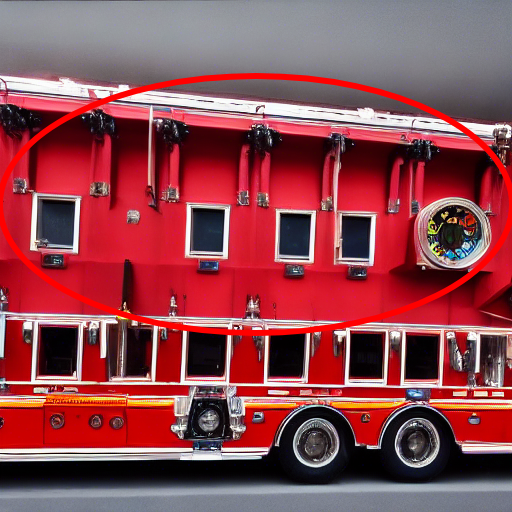} &
      \includegraphics[width=0.13\textwidth]{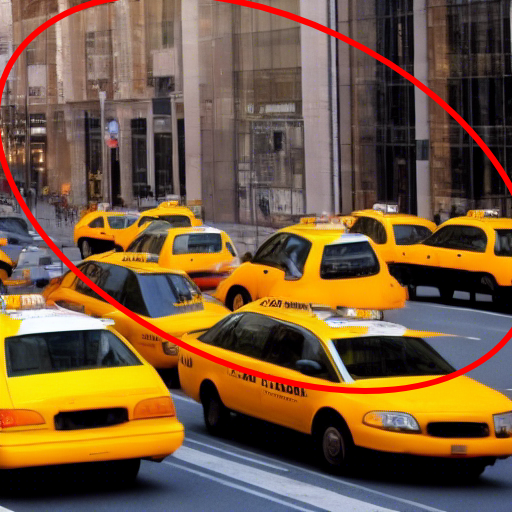} \\
      \raisebox{0.93cm}{\makecell[tt]{\textbf{DDIM} \\ \textit{HURRY}}} & \includegraphics[width=0.13\textwidth]{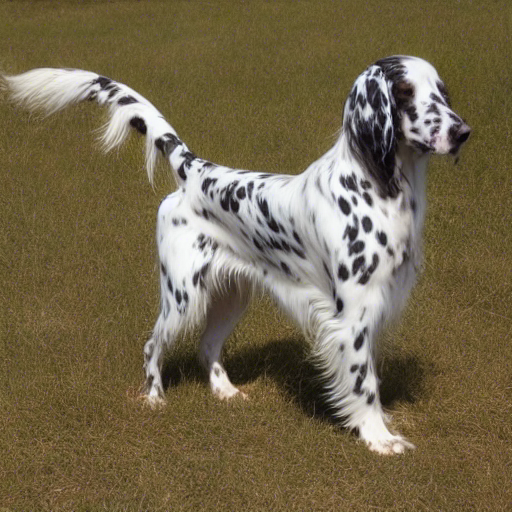} &
      \includegraphics[width=0.13\textwidth]{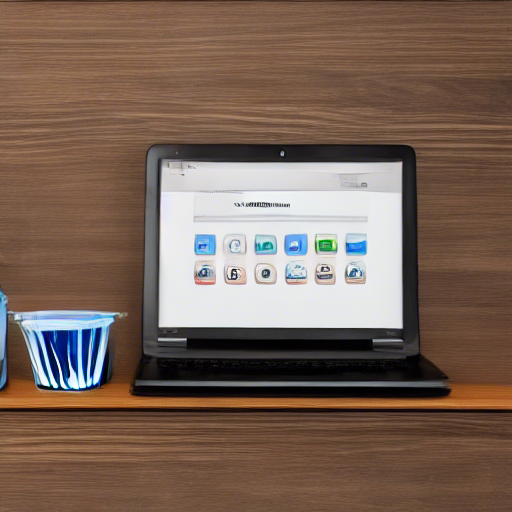} &
      \includegraphics[width=0.13\textwidth]{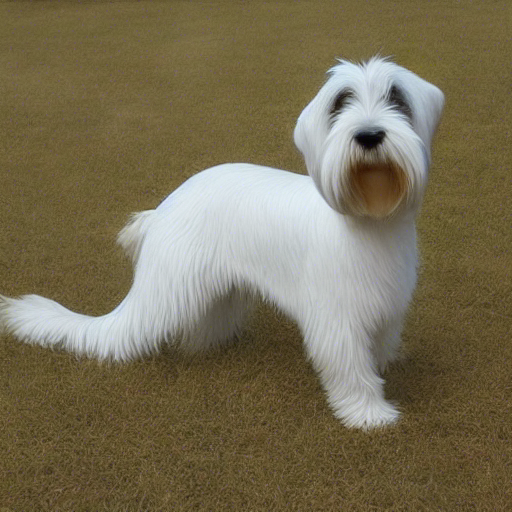} &
      \includegraphics[width=0.13\textwidth]{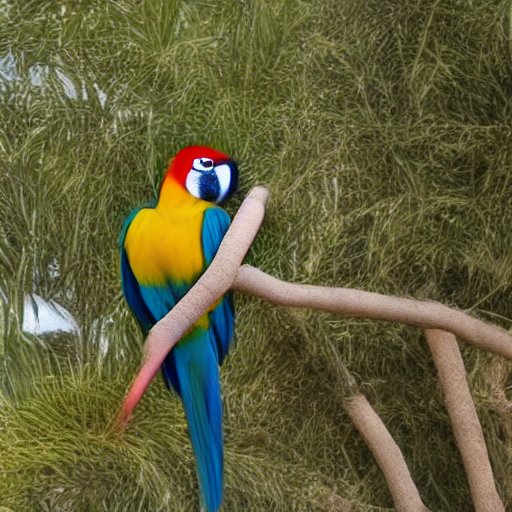} &
      \includegraphics[width=0.13\textwidth]{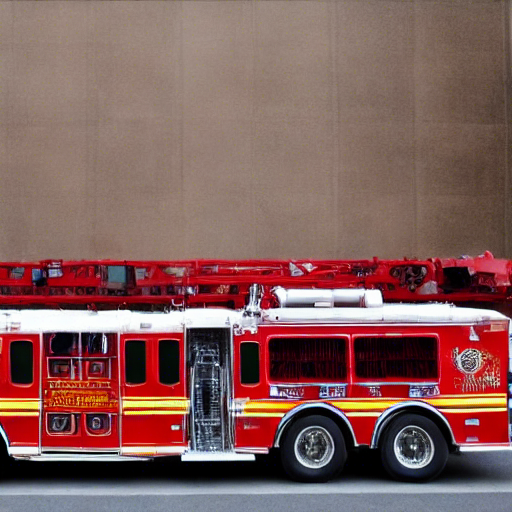} &
      \includegraphics[width=0.13\textwidth]{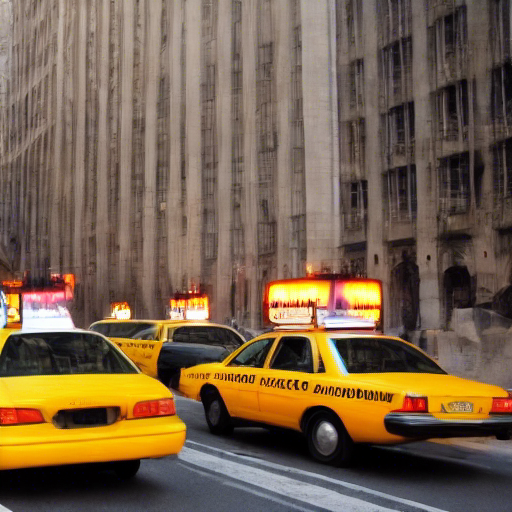} \\
      \raisebox{0.93cm}{\makecell[tt]{\textbf{DDIM} \\ \textit{PHAST}}} & \includegraphics[width=0.13\textwidth]{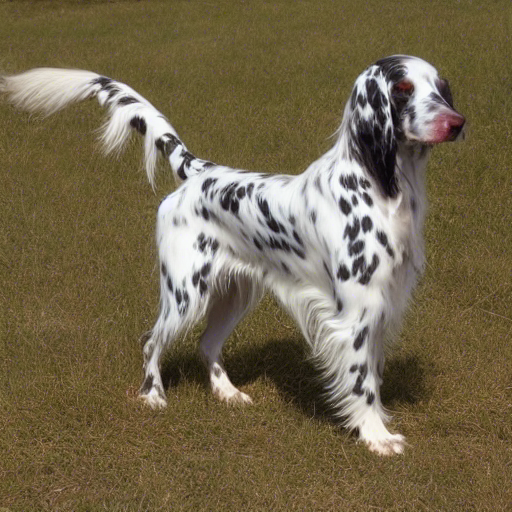} &
      \includegraphics[width=0.13\textwidth]{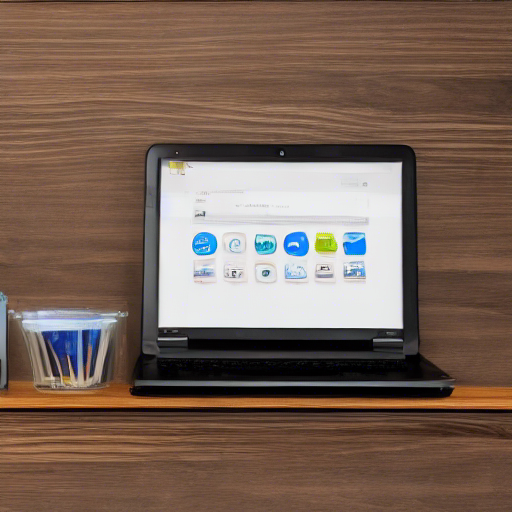} &
      \includegraphics[width=0.13\textwidth]{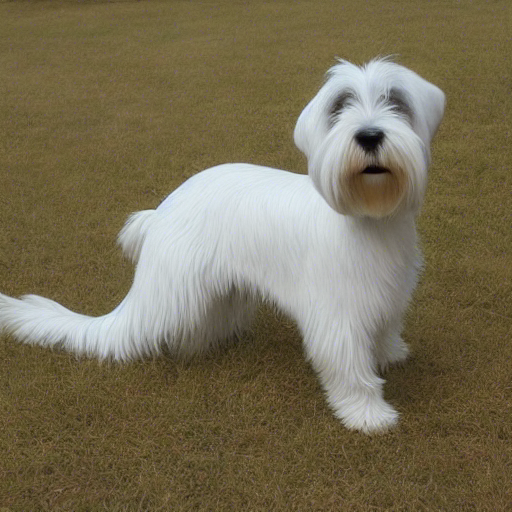} &
      \includegraphics[width=0.13\textwidth]{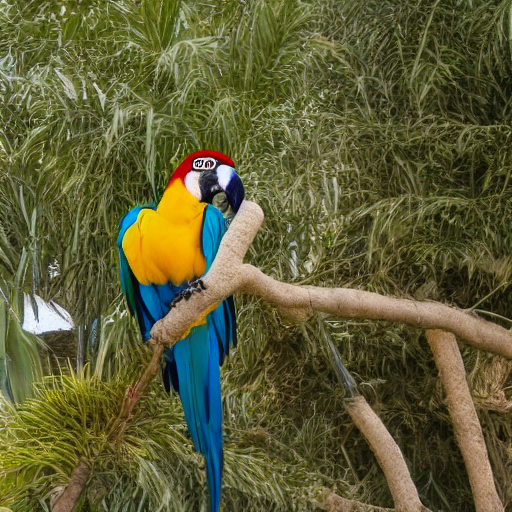} &
      \includegraphics[width=0.13\textwidth]{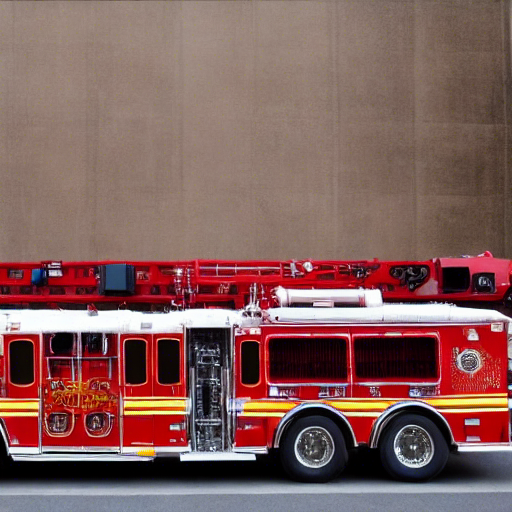} &
      \includegraphics[width=0.13\textwidth]{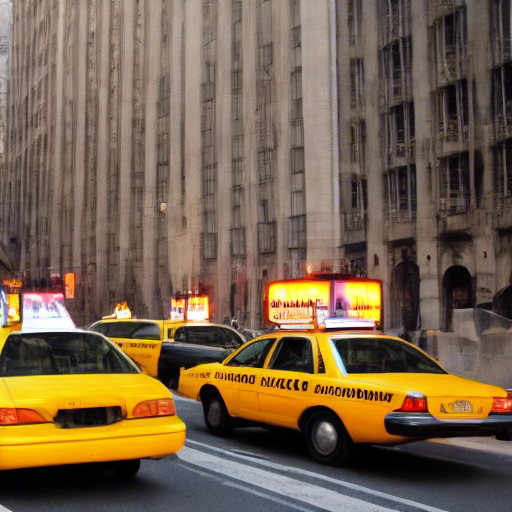} \\
    \end{tabular}
  \end{minipage}
  \caption{Visual comparison of the different sampling techniques for \deleted{the}{ImageNet prompts} (column-wise): `English Setter'; `laptop, laptop computer'; `Sealyham terrier Sealyham'; `macaw'; `fire engine, fire truck'; `cab, hack, taxi, taxicab'. \added{Red circles have been added to the images on the bottom row to help readers identify some of the key differences between the 20-step and 13-step DDIM sampler.} The samplers are (row-wise): 20-step DDIM; 13-step DDIM; HURRY with 10 reuse steps; PHAST with 10 reuse steps. The final three samplers all have a latency of $\sim$4000ms.}
  \vspace{-0.3cm}
  \label{fig:Vis}
\end{figure*}

\vspace{0.5\baselineskip}
\noindent
\textbf{Initial Comparison Between Step-Reduction and Reuse.}
We have now demonstrated that HURRY and PHAST are promising reuse strategies. 
But do they outperform step-reduction, the canonical approach to reducing latency?
Figure \ref{fig:Results} compares step-reduced DDIM samplers with reuse strategies acting on a full 20-step DDIM sampler, for latencies between 3500ms and 5500ms.
The results demonstrate that our reuse strategies consistently outperform step-reduced samplers at comparable latencies.
Figure \ref{fig:Vis} visually supports these results with samples taken from a cross-section of Figure \ref{fig:Results}, at a latency of $\sim$4000ms.
The samples generated by the reuse strategies are more similar to those produced by the original sampler than their step-reduced counterparts.
Notably, the step-reduced samplers produce distorted images --- such as the firetruck which looks more like a building, the detached tail of the Terrier, or the distorted macaw.

\vspace{0.5\baselineskip}
\noindent
The latencies in Figure \ref{fig:Results} are calculated by appropriately summing two components: (1) the latency of a full U-Net call; and (2) the latency of a reuse U-Net call. 
These latencies are estimated by calculating the share of total clock cycles used by each block of the U-Net and scaling the total latency accordingly.
We establish that a full U-Net call has an approximate latency of 152ms, and reuse has an approximate latency of 47ms.\footnote{We derive this latency under the assumption that reuse can reduce the latency of attention blocks by 90$\%$ - a conservative estimate.} 
For each number of reuse steps (i.e., each latency) in Figure \ref{fig:Results}, PHAST is separately selected by algorithm \ref{alg:PHAST}.

\vspace{0.5\baselineskip}
\noindent
\textbf{Further Comparisons Between Step-Reduction and Reuse.}
So far, this paper has focused (almost solely) on the PSNR of modifications to a 20-step DDIM sampler acting on Stable Diffusion (SD) v1.5 with ImageNet labels as prompts. 
We must now demonstrate that our reuse strategies also excel with different datasets, models, sampling procedures, and measures of performance.
Table \ref{tab:SM} starts this wider evaluation by taking the PSNR and FID for SD v1.5 on numerous datasets.
Additionally, it evaluates PHAST on a more powerful model (SDXL) with a complex dataset (PartiPrompts). 
For this table, and all datasets in this section, we \deleted{don't}\added{do not} re-run algorithm \ref{alg:PHAST}. 
As the underlying dynamics of the model are independent of \deleted{its}\added{the} prompt\deleted{s}, we reuse the same strategy (PHAST) that was selected by algorithm \ref{alg:PHAST} given ImageNet labels as prompts.

\vspace{0.5\baselineskip}
\noindent
For every dataset in Table \ref{tab:SM}, PHAST significantly outperforms step-reduction w.r.t. PSNR.
However, they are relatively indistinguishable w.r.t. FID, which suggests that these approaches generate images that are equally coherent and realistic.
Yet, on SDXL with PartiPrompts (See Fig.~\ref{fig:1}), the step-reduced sampler generates a bowl of Pho with an atypical sub-bowl and an impossibly contorted tree that lacks a reflection.
As such, we suggest that the FID might not faithfully track whether or not the image is realistic.
PHAST and HURRY avoid these unrealistic distortions, as they are designed to maximise fidelity w.r.t. the original model\deleted{, which was trained to minimise FID}.

\begin{table}[h]
    \centering
    \resizebox{0.85\columnwidth}{!}{
    \setlength{\tabcolsep}{3pt}
    \begin{tabular}{llccc}
        \toprule
        \textbf{Model}          & \textbf{Dataset}                                  & \textbf{Method}   & \textbf{PSNR} ($\uparrow$) & \textbf{FID} ($\downarrow$)   \\ \midrule
        \multirow{8}{*}{SD1.5}  & MS-COCO 2017                                      & Base(13)          & 17.50         & \textbf{28.22}         \\
                                & {\scriptsize (5k val., 512x512, See Tab.~\ref{tab:FID})}    & PHAST             & \textbf{25.90}         & 29.18         \\ \cmidrule(rl){2-5}
                                & MS-COCO 2014                                      & Base(13)          & 17.53         & 18.71         \\
                                & {\scriptsize (40k val., 512x512)}                 & PHAST             & \textbf{26.03}         & \textbf{17.62}         \\ \cmidrule(rl){2-5}
                                & Instruct-Pix2Pix                                  & Base(13)          & 17.83         & \textbf{59.57}         \\
                                & {\scriptsize (30k val., 512x512)}                 & PHAST             & \textbf{24.98}         & 60.08         \\  \midrule
        \multirow{2}{*}{SDXL}   & PartiPrompts                                      & Base(13)          & 20.30         & 202.58              \\
                                & {\scriptsize (100 random, 1024x1024)}             & PHAST             & \textbf{29.92}         & \textbf{199.88}              \\
        \bottomrule
    \end{tabular}}
    \caption{A comparison of PHAST with a step-reduced DPM++ sampler for various datasets and generative models. The same PHAST strategy, searched for on ImageNet with SD1.5, was used for all of these evaluations. }
    \label{tab:SM}
\end{table}

\begin{table*}[!t]
\centering
\resizebox{\textwidth}{!}{%
\begin{tabular}{l l c c c c c}
\toprule
\textbf{Method} 
  & \textbf{Sampler} 
  & \textbf{PSNR} ($\uparrow$) 
  & \textbf{CLIP} ($\uparrow$) 
  & \textbf{FID} ($\downarrow$) 
  & \textbf{Lat.} (ms)$^\dagger$
  & \textbf{Mem.} (MiB)$^\ddagger$ \\
\midrule
\multicolumn{7}{c}{\textbf{Stable Diffusion v1.5}} \\
\midrule
\multirow{2}{*}{Base(20)}
  & DDIM   & \textit{ref.} & 0.2671 & 28.23 
          & \multirow{2}{*}{4052} & \multirow{2}{*}{8091} \\
  & DPM++  & \textit{ref.} & 0.2673 & \textbf{27.11}
          &                &                \\
\midrule
\multirow{2}{*}{Base(13)}
  & DDIM   & 17.50         & \textbf{0.2681} & 28.22
          & \multirow{2}{*}{2664} & \multirow{2}{*}{8091} \\
  & DPM++  & 17.16         & 0.2676 & 27.39
          &                &                \\
\midrule
\multirow{2}{*}{PHAST}
  & DDIM   & 25.90         & 0.2667 & 29.18 
          & \multirow{2}{*}{2937} & \multirow{2}{*}{12157} \\
  & DPM++  & 23.02         & 0.2664 & 27.95
          &                &                \\
\midrule
\multirow{2}{*}{PHAST$_{\text{FP16}}$}
  & DDIM   & \textbf{25.95} & 0.2667 & 29.21
          & \multirow{2}{*}{3164} & \multirow{2}{*}{10141} \\
  & DPM++  & 23.03 & 0.2664 & 27.95
          &                &                \\
\midrule
\multicolumn{7}{c}{\textbf{Stable Diffusion v1.5 + CFG Dist.} $^{\dagger\dagger}$} \\
\midrule
\multirow{2}{*}{Base(20)} 
  & DDIM   & \textit{ref.} & 0.2611 & 31.71
          & \multirow{2}{*}{2090} & \multirow{2}{*}{7273} \\
  & DPM++  & \textit{ref.} & 0.2613 & 31.66 
          &                &                \\
\midrule
\multirow{2}{*}{Base(13)} 
  & DDIM   & 17.68         & \textbf{0.2625} & 31.96
          & \multirow{2}{*}{1382} & \multirow{2}{*}{7273} \\
  & DPM++  & 16.93         & 0.2619 & \textbf{31.16}
          &                &                \\
\midrule
\multirow{2}{*}{PHAST} 
  & DDIM   & \textbf{28.63} & 0.2614 & 32.56
          & \multirow{2}{*}{1517} & \multirow{2}{*}{9093} \\
  & DPM++  & 25.34 & 0.2612 & 32.66
          &                &                \\
\midrule
\multirow{2}{*}{PHAST$_{\text{FP16}}$}
  & DDIM   & \textbf{28.63} & 0.2614 & 32.57
          & \multirow{2}{*}{1623} & \multirow{2}{*}{8045} \\
  & DPM++  & 25.34 & 0.2612 & 32.66
          &                &                \\
\bottomrule
\end{tabular}}
\resizebox{0.98\columnwidth}{!}{$^{\dagger}$Avg. time needed to generate one image when running continuously for 5 minutes on an NVidia A40 (46GB) with maxed-out batch size (DPM++, full precision).}
\resizebox{0.9\columnwidth}{!}{$^{\ddagger}$Total memory needed to generate one image (DPM++, full precision). $\quad ^{\dagger\dagger}$Our reimplementation of "stage one distillation" from \cite{meng2023distillation}.}
\caption{A comparison of our best strategy (PHAST) with a base sampler (DPM++ and DDIM) on the MS-COCO 2017 validation set, looking at PSNR, CLIP-Score, and FID. We evaluate our samplers on Stable Diffusion v1.5 and a low-latency distilled alternative that fuses the conditional and unconditional forward passes into one. \added{We also include measurements of latency and memory for reference.}}
\label{tab:FID}
\end{table*}

\vspace{0.5\baselineskip}
\noindent
In Table \ref{tab:FID}, we evaluate PHAST against step-reduction for two samplers (DDIM and DPM++) on the MS-COCO \cite{lin2014microsoft} validation dataset.
Algorithm \ref{alg:PHAST} was ran twice to separately select PHAST for each sampler; however, we found that up to a timestep rounding error, these two strategies were the same.
In particular, by directly comparing the timesteps (1-1000) rather than the steps (1-20), any differences between the two reuse strategies reduced to a rounding error.
This suggests that our search algorithm can generalise across different samplers, which is perhaps unsurprising as the U-Net is unchanged between samplers. 

\vspace{0.5\baselineskip}
\noindent
Table \ref{tab:FID} shows that PHAST significantly outperforms step-reduction for both samplers w.r.t. PSNR, yet it is marginally worse for CLIP and FID.
But\added{,} as \deleted{we've}\added{we have} previously alluded to, these more subjective measures of performance might not be completely faithful (See Figs.~\ref{fig:1} and \ref{fig:Vis}).
Interestingly, the PSNR of PHAST is consistently larger for DDIM over DPM++.
We attribute this difference to the more linear nature of DDIM, which might aid reuse.

\vspace{0.5\baselineskip}
\noindent
Along with evaluating PHAST for two different samplers, Table \ref{tab:FID} also considers two different models.
The top row\added{s} \deleted{is}\added{are} a vanilla version of Stable Diffusion, and the bottom row\added{s} evaluate\deleted{s} a model whose conditional and unconditional forward passes of the classifier-free guidance (CFG) are distilled into one, following \cite{meng2023distillation}.
In particular, we distilled the model using the MS-COCO 2013 training dataset with 4 A100 for 2-3 days.\footnote{Since Meng et al. \deleted{didn't}\added{did not} release their code or checkpoints, we had to re-implement their method.
As \deleted{we're}\added{we are} not interested in improving raw performance of DPMs but to approximate them efficiently, we \deleted{didn't}\added{did not} perform a very exhaustive distillation.}
We observe that even when a model has been additionally optimised, the same PHAST strategy --- searched for on vanilla stable diffusion --- results in very similar behaviour, with a high fidelity (PSNR) and slight reduction in FID and CLIP.
This demonstrates that our reuse strategies can be used in tandem with other approaches for optimising a DPM's latency.

\vspace{0.5\baselineskip}
\noindent
In summary, both of the reuse strategies proposed in this paper appear to be optimal or near-optimal.
Moreover, they consistency and significantly outperform step-reduced samplers w.r.t. PSNR, and remain competitive for the more subjective measures of image quality. 
For a given number of reuse steps, we have shown that the strategy selected using SD v1.5 on ImageNet with a DDIM sampler can generalise across models, datasets, and sampling procedures. 

\section{Discussion}\label{sec6}
\textbf{The Memory-Latency Trade-Off.}
The reductions in latency achieved by our reuse strategies come at the expense of increased memory usage, which is required to cache the attention maps for reuse.
Although Table \ref{tab:FID} shows that speedup is not affected by this extra cost, the possibility for utilising reuse might be limited in memory-constrained systems.
To alleviate this, developers could \deleted{be to} opt for caching attention maps in reduced precision.
For instance, we observe that storing attention maps in FP16 does not degrade our results (See Tab.~\ref{tab:FID}), but allows us to halve the memory required to cache attention maps.

\vspace{0.5\baselineskip}
\noindent
We could reduce memory even further by considering 8-bit quantisation --- \added{notably,} PHAST$_{\text{INT8}}$ with our CFG-distilled model and the DPM++ solver achieves an FID of 32.65 and PSNR of 25.35.
However, we also notice that the overhead of performing quantisation becomes a bottleneck in this case, hurting latency\deleted{;}\added{,} \deleted{this}\added{which} is why we did not include \added{a} full comparison.
Having said that, we expect that this might be a feasible way forward for devices that operate on quantised tensors, since the cost of performing quantisation will be amortized in those cases.
We leave further investigation into memory reduction for future work.

\vspace{0.5\baselineskip}
\noindent
\textbf{Refining our Reuse Strategies.}
There are numerous ways to refine the reuse strategies proposed in this paper.
For example, the step-wise strategies ignore a natural axis for search that might bolster performance, namely layer-wise search. 
Rather than employing a reuse vector, we might envision a reuse matrix where the rows index the U-Net's layers and the columns index a step in the sampling procedure. 
We excluded this search from the main body of the paper due to its computational complexity and limited impact on performance for SD v1.5.
However, we acknowledge that for certain DPMs this layer-wise refinement might produce significant improvements on the proposed reuse strategies. 

\vspace{0.5\baselineskip}
\noindent
Additionally, fine-tuning the U-Net to better tolerate reuse might further enhance performance.
This could involve first fixing a reuse strategy, then fine-tuning the model whilst applying this strategy.
Alternatively, the U-Net could be preemptively fine-tuned to better accommodate reuse strategies in general.
For example, a developer might fine-tune the U-Net with random reuse strategies or modify it to have a more linear score function.
We \deleted{didn't}\added{did not} investigate this in the paper, as our focus was to create a training-free method for reducing the cost of calling the U-Net --- an area that has so far been \added{relatively} overlooked. 

\vspace{0.5\baselineskip}
\noindent
\added{A more fundamental adjustment to our paper's attention-focused reuse strategies could involve reusing tensors other than attention maps during a DPM's sampling procedure. 
For example, rather than reusing attention maps, a low-latency DPM sampler could reuse features --- such as the outputs of attention blocks or linear layers --- instead. 
In Table \ref{feature}, we compare the PSNR of reuse strategies that have the same number of reuse steps, with one set reusing features and the other reusing attention maps. 
The results indicate that feature reuse is inferior to attention map reuse for strategies with the same number of reuse steps,  likely because features are derived from both attention maps and other time-dependent components.
Nevertheless, the relatively small difference in PSNR between these approaches, combined with the potential for reduced memory and latency with feature reuse, suggests that future research into the trade-offs between these approaches could be insightful.}

\begin{table}[h]
\centering
\resizebox{0.9\textwidth}{!}{%
\begin{tabular}{l l c c c c}
\toprule
\multirow{2}{*}{\textbf{Metric}} & \multirow{2}{*}{\textbf{Reusing}} & \multicolumn{2}{c}{\textbf{DDIM}} & \multicolumn{2}{c}{\textbf{DPM++}} \\
\cmidrule(lr){3-4}\cmidrule(lr){5-6}
 &  & HURRY & PHAST$^\dagger$ & HURRY & PHAST$^\dagger$ \\
\midrule
\multirow{2}{*}{PSNR $(\uparrow)$} 
 & Attention Maps & \textbf{24.56} & \textbf{27.10} & \textbf{22.49} & \textbf{24.41} \\
 & Features$^\ddagger$   & 22.58 & 25.05 & 20.21 & 21.85 \\
\bottomrule
\end{tabular}}
\resizebox{\columnwidth}{!}{$^{\dagger}$Here, PHAST is the strategy located using algorithm \ref{alg:PHAST} for attention map reuse. \quad $^\ddagger$ We define features as the output of attention blocks.}
\caption{\added{A comparison of PSNR (averaged over 200 ImageNet prompts as labels) for attention map and feature reuse strategies --- HURRY and PHAST --- for a 20-step DDIM and DPM++ sampler with 10 reuse steps.}}
\label{feature}
\end{table}

\vspace{0.5\baselineskip}
\noindent
\textbf{Selecting the Parameters for Reuse.}
This paper has explored the following problem: Given an N-step sampling procedure with $r$ reuse steps, how should these reuse steps be allocated in order to maximise the performance of the model?
In reality, N and $r$ are not necessarily fixed; they could be selected alongside their corresponding strategy, $\boldsymbol{\pi}^{\text{(N,r)}}$, to maximise the model's performance while keeping the latency below a certain threshold. 
However, considerable amounts of computation would be required to solve this constrained optimisation problem.
As such, we leave it for developers to select sensible parameters (N,r) for their specific applications, perhaps via a short process of trail and error with HURRY$^{\text{(N,r)}}$.

\section{Conclusion}\label{sec7}
This paper introduces two reuse strategies that reduce a DPM's latency while retaining the original DPM's weights and number of calls to the U-Net.
We started by analysing the dynamics of the reverse diffusion process, which pointed to a `later-is-better' reuse strategy.
By conducting a local search around this heuristic strategy, we improved the model's performance for a fixed latency.
Both strategies outperformed naive step reduction, especially when remaining faithful to the model's baseline behaviour is of primary concern.
Moreover, we showed that reuse strategies can generalise across models, sampling procedures, and datasets. 
We hope that future work can further investigate redundancies in the reverse diffusion process and their potential for improving a DPM's latency. 

\vspace{0.5\baselineskip}
\noindent
\textbf{Broader Impact.}
Our work reduces the latency of high-quality DPM image synthesis.
While this may pose societal benefits, DPMs can also be used to produce biased or harmful content \cite{anderljung2023protecting}.
Reductions in a DPM's latency might increase the ease with which malicious actors can produce harmful content. 
\backmatter

\section*{Declarations}
\begin{itemize}
\item Funding: This work was partially supported \added{by} Samsung AI Centre Cambridge and SCRTP facilities at the University of Warwick. 
\item Conflict of interest: The authors declare that they have no conflict of interest.
\item Ethics approval and consent to participate: N/A
\item Consent for publication: Consent for publication has been obtained.
\item Data availability: Only public datasets were used in this work.
\item Materials availability: N/A
\item Author contribution: Rosco Hunter and {L}ukasz Dudiziak contributed equally. 
\end{itemize}
\noindent

\bibliography{sn-bibliography}

\end{document}